\documentclass[times, review, 10pt]{elsarticle}

\usepackage{amssymb}
\usepackage{amsmath}
\usepackage{multirow}
\usepackage{booktabs}

\usepackage{wrapfig}
\usepackage{graphicx}

\usepackage{color} 
\usepackage{xcolor}
\newcommand{\modulename}{GCR}

\newcommand{\eg}{\textit{e}.\textit{g}.}

\usepackage{graphicx}%
\usepackage{multirow}%
\usepackage{amsmath,amssymb,amsfonts}%
\usepackage{amsthm}%
\usepackage{mathrsfs}%
\usepackage[title]{appendix}%
\usepackage{xcolor}%
\usepackage{textcomp}%
\usepackage{manyfoot}%
\usepackage{booktabs}%
\usepackage{algorithm}%
\usepackage{algorithmicx}%
\usepackage{algpseudocode}%
\usepackage{listings}%

\usepackage{colortbl}
\definecolor{mygray}{gray}{.9}

\journal{Pattern Recognition}

\begin{document}

\begin{frontmatter}



\title{ClickTrack: Towards Real-time Interactive Single Object Tracking} 

\author[university]{Kuiran Wang\fnref{equal}}
\ead{wangkuiran19@mails.ucas.ac.cn}
\author[university]{Xuehui Yu\fnref{equal}}
\ead{yuxuehui17@mails.ucas.ac.cn}
\author[university]{Wenwen Yu}
\ead{yuwenwen22@mails.ucas.ac.cn}
\author[university]{Guorong Li}
\ead{liguorong@ucas.ac.cn}
\author[institute]{Xiangyuan Lan}
\ead{xiangyuanlan@life.hkbu.edu.hk}
\author[university]{Qixiang Ye}
\ead{qxye@ucas.ac.cn}
\author[university]{Jianbin Jiao}
\ead{jiaojb@ucas.ac.cn}
\author[university]{Zhenjun Han\corref{corresponding}}
\ead{hanzhj@ucas.ac.cn} 

\cortext[corresponding]{Corresponding author.}
\fntext[equal]{The two authors contribute equally to this work.}

\affiliation[university]{organization={School of Electronic, Electrical and Communication Engineering, University
of Chinese Academic of Sciences (UCAS)},
            addressline={}, 
            city={Beijing},
            postcode={101480}, 
            state={},
            country={China}
            }
\affiliation[institute]{
organization={Institute of Vision Intelligence, Peng Cheng Laboratory},
addressline={No. 2, Xingke 1st Street},
city={Shenzhen},
postcode={518000},
state={Guangdong Province},
country={China}
}

\begin{abstract}
Single object tracking(SOT) relies on precise object bounding box initialization. In this paper, we reconsidered the deficiencies in the current approaches to initializing single object trackers and propose a new paradigm for single object tracking algorithms, ClickTrack, a new paradigm using clicking interaction for real-time scenarios. Moreover, click as an input type inherently lack hierarchical information. To address ambiguity in certain special scenarios, we designed the Guided Click Refiner(GCR), which accepts point and optional textual information as inputs, transforming the point into the bounding box expected by the operator. The bounding box will be used as input of single object trackers. Experiments on LaSOT and GOT-10k benchmarks show that tracker combined with GCR achieves stable performance in real-time interactive scenarios. Furthermore, we explored the integration of GCR into the Segment Anything model(SAM), significantly reducing ambiguity issues when SAM receives point inputs.
\end{abstract}




\begin{keyword}
Click; Single Object Tracking; Video Object Segmentation; Real-time Interaction
\end{keyword}

\end{frontmatter}

\section{Introduction}
\label{intro}
\begin{figure}[htbp]
\centering
\includegraphics[width=0.7\linewidth]{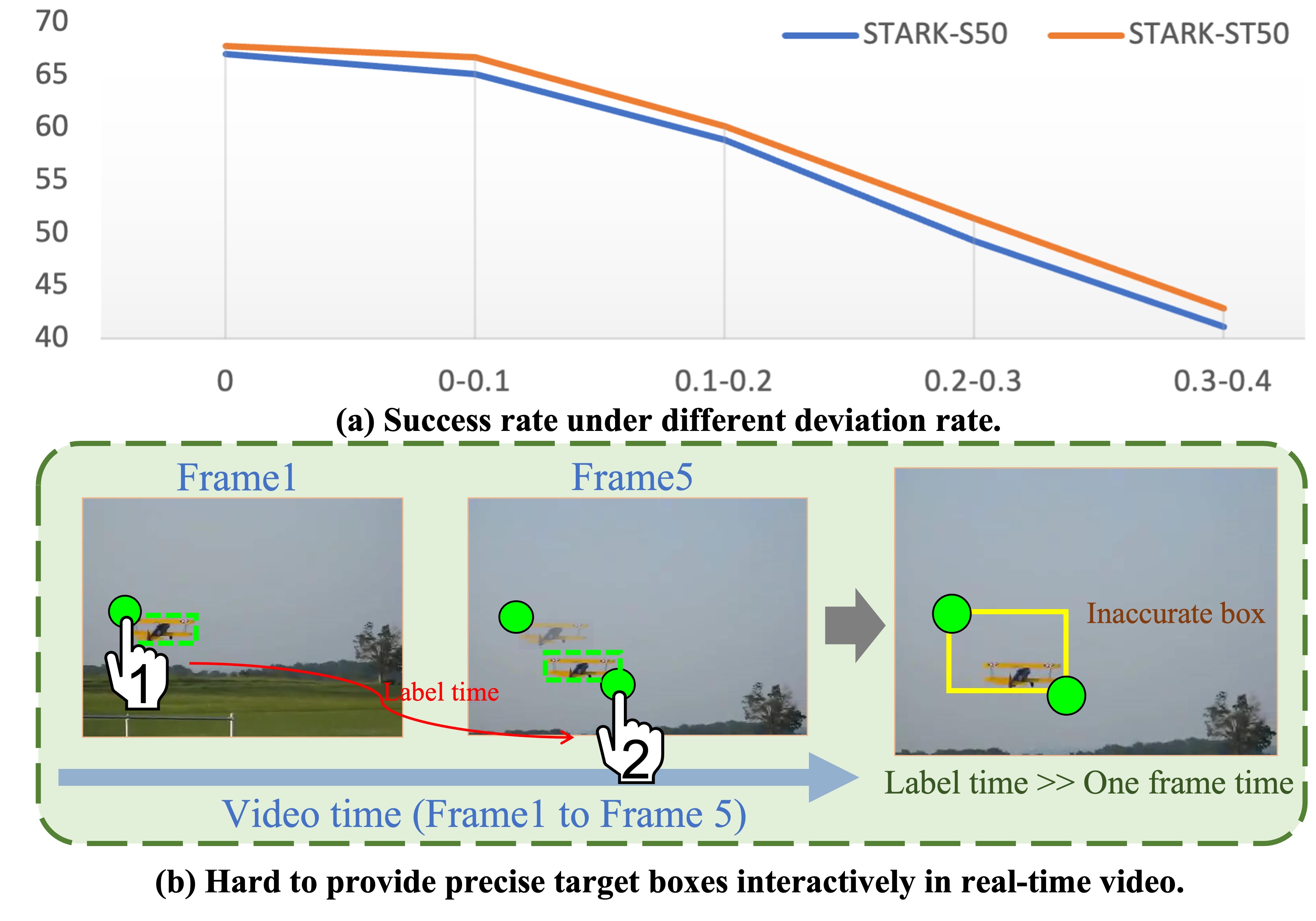}
\caption{(a) The performance drops when the deviation rate of annotated bounding box increases. (b) The \textcolor{green}{green} dashed box means the object position in the current frame (starting annotating from the left-top of the object) and the \textcolor{yellow}{yellow} box means the annotated box in the $k$-th frame (finishing the bounding box annotation at the right-down of the object), the annotated bounding box is inaccurate. }

\label{fig:performance_drop}
\end{figure}

Traditional single object tracking (SOT) aims to locate and track an arbitrary object throughout a video sequence. Recent advancements in deep learning have significantly improved tracking accuracy and success rate. However, SOT trackers heavily rely on precise initial bounding box annotations in the first frame to establish the template feature of the tracking object. Any inaccuracies in the annotation process can adversely affect tracking accuracy, as shown in Fig.~\ref{fig:performance_drop} (a), an increase in the deviation rate of the initial bounding box leads to a sharp drop in tracking performance (success rate). Thus, the accuracy of annotation plays a crucial role in these trackers.

However, annotating the tracked object within the video stream in real-time interactive scenarios introduces additional challenges. As the object's position continually changes across different frames during manual annotation, there is a higher likelihood of obtaining an inaccurate bounding box, as illustrated in Fig.~\ref{fig:performance_drop} (b). The example images are from the LaSOT~\cite{fan2019lasot} benchmark, with an interval of 3 frames. It is obvious that even completing the annotation of the initial box within five frames may impact tracking accuracy. 

\begin{figure}[htbp]
\centering
\includegraphics[width=0.99\linewidth]{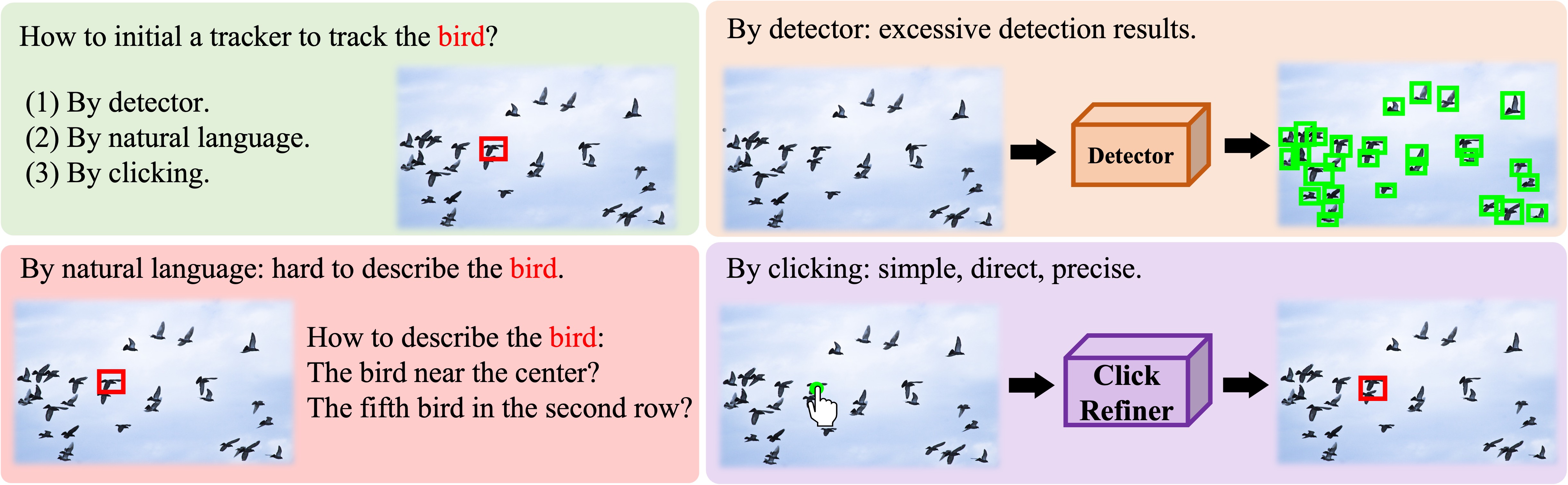}
\caption{Different initialization methods for single object tracker.}
\label{fig:different initilization method.}
\end{figure}

The effectiveness of single object trackers largely depends on the choice of a proper initialization method, making it crucial for practical applications. We reevaluated different initialization methods for the single object tracking task in real-time interactive scenarios. As shown in Fig.~\ref{fig:different initilization method.}, the first initialization method uses a detector. This method has several issues: (1) the detector outputs multiple bounding boxes, and further interaction is required to accurately locate the target to be tracked; (2) if the detector fails to detect a target, it cannot specify a target in the current frame. Thus, detector-based initialization has significant limitations in real-time interactive scenarios. The second method, introduced in natural language tracking tasks~\cite{DBLP:conf/wacv/FengABLS20, DBLP:conf/cvpr/WangSZJW0W21, DBLP:conf/cvpr/LiYCP22, DBLP:journals/corr/abs-2303-12027, DBLP:conf/cvpr/FengABS21}, involves specifying the tracking target via a natural language description. The main problems with this method are: (1) precisely specifying a tracking target requires an accurate natural language description, which increases the thinking time cost of interaction and decreases stability; (2) precisely locating a specific target using only a natural language description is challenging in scenarios containing many similar targets. Considering these limitations, designing a more streamlined and stable initialization method for real-time interactive scenarios is highly meaningful.

\begin{figure}[htbp]
\centering
\includegraphics[width=0.65\linewidth]{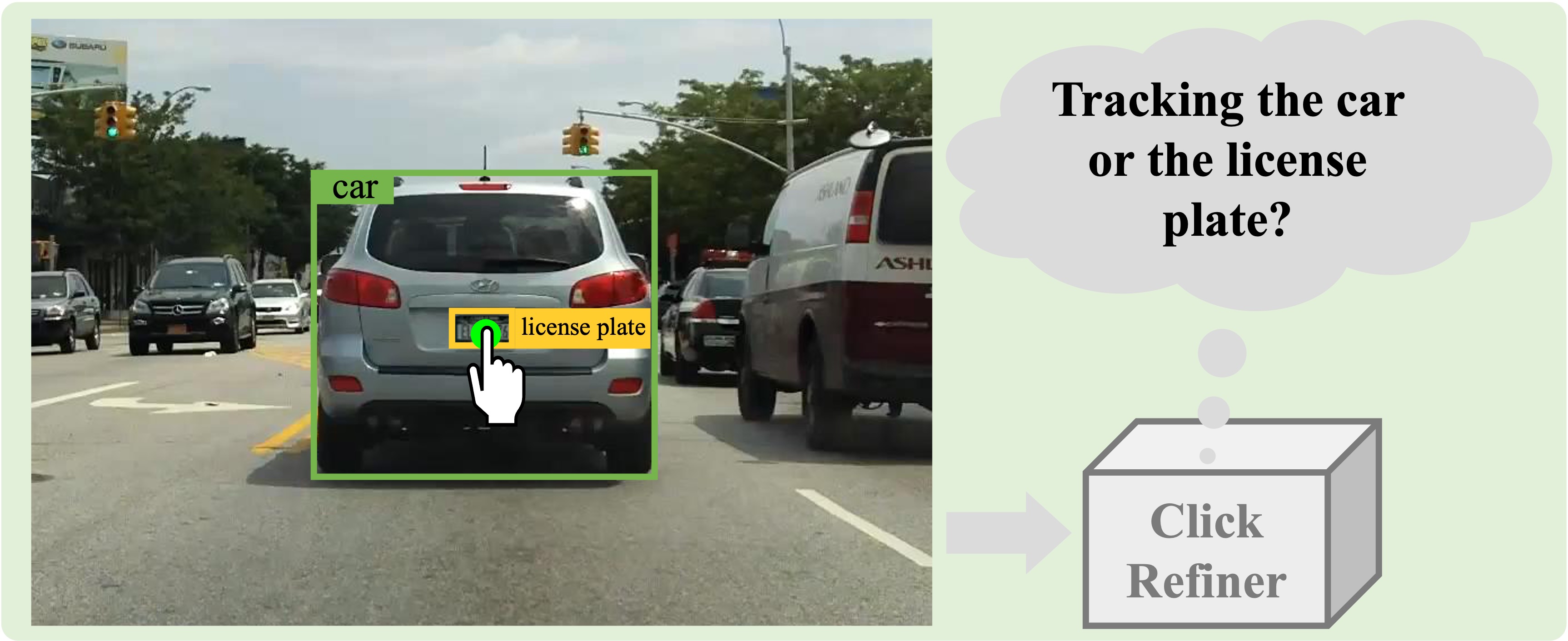}
\caption{Tracking ambiguity: when the point is clicked on the overlap region, \eg, the green point on the license plate, the tracker is confused about which one is for tracking.}
\label{fig:confusion}
\end{figure}

Currently, the clicking interaction method~\cite{Click,chen2022point,kirillov2023segment,DBLP:Attnshift,yu2022object} has garnered widespread attention from researchers. We believe that clicking, because of its speed, simplicity and precision, is particularly well-suited as an initialization method for single-object trackers in real-time interactive scenarios. Although clicking has many advantages, it lacks hierarchical information.
The absence of hierarchical information often compromises the model’s precision in localizing the target within certain scenarios. As illustrated in Fig.~\ref{fig:confusion}, when a click is made on a license plate, the model cannot determine whether the operator intends to track the license plate or the entire car. We suggest that in such scenarios, introducing simple category information could help the model determine the specific area to be tracked. It is important to note that, unlike natural language tracking tasks, our approach only introduces simple category information, such as ``car" or ``license plate", without descriptions of target location or appearance. Consequently, in scenarios where category information is necessary, operators can efficiently communicate the category via voice with minimal cognitive effort. In ordinary scenarios, a simple click operation is all that is needed.

Building on the considerations outlined above, a Guided Convolution (GC) structure was initially designed to accommodate both visual and guiding features.
The guiding features may be simple category information derived from text generated by a natural language model, or they can be learnable features. The structure of the guided convolution allows for interaction between the guiding features and visual features, resulting in the generation of predicted bounding boxes that align with the operator's expectations. Based on GC, we developed two additional modules: Prototype Selection (PS) and Iterative Refinement (IR), which collectively form the Guided Click Refiner (GCR) regression model. In summary, the point provides the precise location of the tracked object, while the guided feature (textual information or learnable feature) guides the generation of the bounding box as the operator expects.
In this paper, the GCR is conceptualized as a standalone model, crafted to provide initialization for single-object trackers in real-time interactive scenarios.
This design allows the GCR to be integrated with any single-object tracker, meeting the needs of real-time interactive scenarios. We have named this initialization approach for single-object tracking as ClickTrack.

We conducted extensive experiments on the LaSOT~\cite{fan2019lasot} and GOT-10k~\cite{huang2019got} datasets combining GCR with transformer-based tracker STARK~\cite{yan2021learning}. The results demonstrate that the GCR model achieves good tracking accuracy with a single point and text input, which is close to the performance of precise initial annotations. Furthermore, experimental analysis reveals the robustness of our GCR model to single-point location inputs, with processing speeds that satisfy real-time interaction demands. Moreover, visualization results indicate that our method incorporating text information effectively alleviates the ambiguity caused by single-point annotations.

Additionally, the recently proposed Segment Anything Model(SAM)~\cite{kirillov2023segment} can accept points as input, enabling seamless integration with any video object segmentation(VOS) methods~\cite{khoreva2019lucid,yan2023universal,zhang2020transductive,robinson2020learning} to address the requirements of real-time interactive scenarios. However, SAM also faces ambiguity issues arising from single-point input. Therefore, we integrated the GCR model structure into SAM, conducting exploratory experiments. The results demonstrate that the GCR structure effectively mitigates the ambiguity problem caused by single-point input and enhances SAM's performance in real-time interactive scenarios. 

We anticipate the promising results and comprehensive analysis presented in this work will draw the research community's attention to the new SOT paradigm: ClickTrack. The key contributions of this paper can be summarized as follows:
\begin{itemize}
    \item We reconsidered the initialization method for single-object trackers in real-time interactive scenarios and propose a new paradigm for SOT: ClickTrack.
    \item We proposed a novel architecture for real-time interactive tracking which takes point and guided information (text feature or learnable feature) as input to provide a precise bounding box for the tracker. 
    \item Our experimental results demonstrate that in real-time interactive scenarios, compared to other initialization methods, GCR can provide more stable tracking performance through click interaction.
    \item The exploratory experiment of combining \modulename \ 
    with the SAM demonstrates that the GCR model's structure effectively resolves ambiguity issues caused by single-point input and is transferable.
\end{itemize}

\section{Related Work}
\subsection{Natural Language Tracking}
Inspired by the visual grounding
task development, Li et al.~\cite{DBLP:conf/cvpr/LiTGSS17} define the task of tracking by natural language specification. Yang et al.~\cite{DBLP:journals/tcsv/YangKCSL21} decompose the problem into three sub-tasks, i.e., grounding, tracking, and integration, and process each sub-task separably by three modules. Differently, Feng et al.~\cite{DBLP:conf/wacv/FengABLS20} solve this task following
the tracking-by-detection formulation, which utilizes natural language to generate global proposals on each frame for
tracking. To provide a specialized platform for the task of
tracking by natural language specification, Wang et al.~\cite{DBLP:conf/cvpr/WangSZJW0W21}
release a new benchmark for natural language-based tracking named TNL2K and propose two baselines initialized by
natural language and natural language with bounding boxes,
respectively. Li et al.~\cite{DBLP:conf/cvpr/LiYCP22} employ a target-specific retrieval
module to localize the target, which is used to initialize a local tracker. 

Unlike the setting of natural language tracking tasks, we try to determine the tracking target through more direct click interactions and combine it with powerful trackers to achieve better performance in real-time interactive scenarios.

\subsection{Single Object Tracking}
Single object tracker tracks the object specified in the first frame. Compared with traditional correlation filter tracking methods, recent Siamese network-based trackers~\cite{bertinetto2016fully,li2018high,li2019siamrpn++} have made amazing progress in the performance of single object tracking. Li et al. presented SiamRPN~\cite{li2018high} and SiamRPN++~\cite{li2019siamrpn++}, which try to introduce object detection progress into object tracking for more accurate location prediction. Nowadays, many scholars have tried to apply the Transformer~\cite{vaswani2017attention} to the field of object tracking, which improves tracking performance by learning discriminative object representations~\cite{yan2021learning,li2024transformer,gao2024transformer}, and the effect has exceeded the networks based on CNN.

Although these works have achieved significant performance improvements, they all rely on precise initial bounding boxes, which limits their application in real-time interactive scenarios. This paper focuses on the issue of initializing single-object tracking algorithms. It achieves the initialization of single-object tracking algorithms through clickable interactions to specify the tracking target, meeting the needs of real-time interactive scenarios.

\subsection{Point-based Vision Tasks}
Point annotation has recently been studied as an extremely cost-saving labeling method. Compared with accurate bounding box annotations, point annotation is a fairly recent innovation, which is easier to obtain. There have been some researches on point supervision in vision tasks such as object localization~\cite{yu2022object,song2021rethinking}, object detection~\cite{chen2022point,chen2021points}, crowd counting~\cite{tabernik2024dense} and instance segmentation \cite{liew2017regional,maninis2018deep}. Point-based instance segmentation~\cite{li2018interactive,benenson2019large} is usually employed in an interactive manner where the models are trained with full supervision. 
In contrast to the aforementioned work, we use clicking as the initialization method for a single-object tracker in real-time interactive scenarios. By integrating it with single object tracker, the tracker meets the requirements for instantly specifying tracking targets in real-time interactive contexts, thereby establishing a new paradigm for target tracking, ClickTrack.

\subsection{Multimodality Tasks}
Recently, there has been a trend to develop vision-and-language approaches to visual recognition problems, \eg, visual question answering~\cite{lu2019vilbert}, image captioning~\cite{zhou2020unified,ijcai2021p91}. Vision-language pre-training has attracted growing attention during the past few years. As a milestone, Radford et al. devise a large-scale pre-training model, named CLIP~\cite{radford2021learning}, which performs cross-modal contrastive learning on hundreds or thousands of millions of image-text pairs. Currently, CLIP has been tried to be applied in the field of computer vision, including semantic segmentation~\cite{kim2024lc}, object detection~\cite{song2024prompt,li2022grounded}, video field~\cite{luo2021clip4clip}, \emph{etc}. 

By referencing related work in multimodality, we aim to introduce simple category information to enable operators to more precisely specify tracking areas in special scenes through input text information.

\label{subsec1}

\section{Methodology}
In this section, we first review the general setting of single object tracking and introduce the ClickTrack paradigm in detail. Then, we design two simple baseline models with reference to two well-known object detectors, Faster R-CNN~\cite{ren2015faster} and FCOS~\cite{tian2019fcos}. 
Subsequently, we analyze the strengths and limitations of the two simple baseline models and consider the necessity of incorporating text information in some special scenarios to address the ambiguity problem associated with single point input, designing the Guided Click Refiner (GCR) model. The GCR model consists of the Guided Convolution (GC) structure, the prototype selection (PS) module, and the iterative refinement (IR) module.

\begin{figure*}[htbp]
\centering
\includegraphics[width=1.\linewidth]{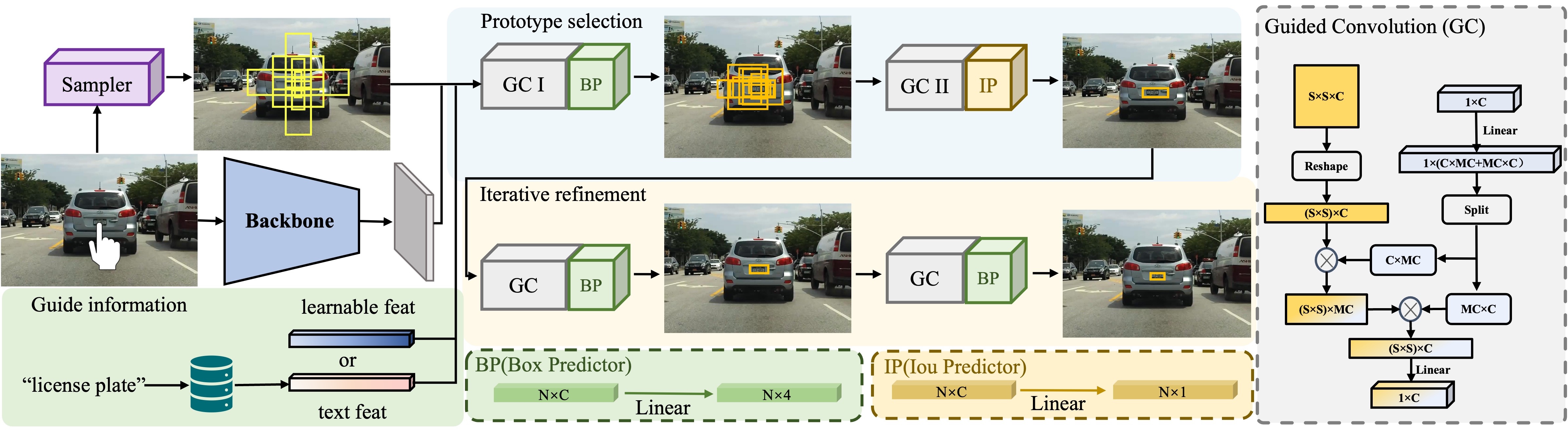}
\caption{The Guided Click Refiner (\modulename) framework, including the Prototype Selection module and Iterative Regression module. With the click annotation and guiding information, the refined bounding box is obtained by \modulename.}
\label{fig:framework}
\end{figure*}

\subsection{Revisiting Single Object Tracking}
Traditional single object tracking (SOT)~\cite{yan2021learning, bertinetto2016fully,li2018high,li2019siamrpn++} typically requires an accurate bounding box annotation for the target object in the initial frame of a video sequence. Given the initial object bounding box, denoted as $b_0$, and the tracking sequence, denoted as $S$, the tracker $\mathcal{T}$ utilizes the input to produce tracking results for subsequent frames. The tracking process can be described as follows:

\begin{equation}\label{equ:traditional}
\mathbf{b}=[b_1,\dots,b_n]=\mathcal{T}(S, b_0),
\end{equation}where $b_i$ denotes the bounding box of the tracked object predicted in the $i$-th frame.

In practical application scenarios, the initialization of $b_0$ is essential for single-object trackers. However, as described in Section \ref{intro}, the various initialization methods currently available exhibit certain limitations in real-time interactive scenarios. We believe clicking is more suitable as the interactive method for real-time initialization of single-object trackers. Consequently, the tracking process is transformed as follows:

Specifically, we denote the point annotation in the initial frame as $p_0$ and introduce a point-to-box regressor, denoted as $\mathcal{R}$. Consequently, the single object tracking paradigm can be transformed as follows:

\begin{equation}\label{equ:traditional}
\mathbf{b}=\mathcal{T}(S, \mathcal{R}({p_0})).
\end{equation}

In this way, no specific modifications to existing trackers are required, and our module can simply connect to them. We name this new SOT paradigm ClickTrack.

\subsection{Na\"ive Point-based Object Discovery}\label{sec:Naive Point-based Object Discovery}
The main challenge of ClickTrack is to discover the whole object from an initial point and use a bounding box to represent the object. Then the estimated bounding box can be viewed as the template for the tracker and track the object in the following frames.

It is obvious that the regression outputs for object bounding boxes in conventional object detectors align well with our proposed design.
Therefore, inspired by classic object detectors Faster R-CNN~\cite{ren2015faster} and FCOS~\cite{tian2019fcos}, we first design two naive point-based regressors to tackle the challenge.

We first define the objective of the regression: the distance from the labeled point to the four edges of the object bounding box.
Specifically, the distance to each edge of the bounding box is predicted from the labeled point $p_0=(p_x, p_y)$. The prediction is denoted as $delta=(\hat{l},\hat{t},\hat{r},\hat{b})$, representing the distances in the left, top, right, and bottom directions, respectively. Then, the point can be converted into a bounding box $b=(x_{0},y_{0},x_{1},y_{1})$:
\begin{equation}
\begin{aligned}\label{equation: convert box}
x_{0}=p_x - \hat{l}, \quad y_{0}=p_y - \hat{t}, \\
x_{1}=p_x + \hat{r}, \quad y_{1}=p_y + \hat{b}.
\end{aligned}
\end{equation}

In this paper, we study how to obtain a high-quality bounding box with point annotation. We define
$I$ as the input image, $\mathcal{F}(I)$ as the backbone features from the image encoder. We consider two simple ways to perform regression prediction: Box Feature Refiner (BFR) and Point Feature Refiner (PFR). 

\indent
\textbf{Box Feature Refiner.}
We extract the RoI feature of $\mathcal {B}$ from $\mathcal{F}(I)$ by RoIAlign $\mathcal{A}(\cdot)$\cite{he2017mask}, where $\mathcal{B}$ is centered at the labeled point $p_0=(p_x, p_y)$, and its dimensions are determined by the shortest distance from this point to the image boundary.

Finally, the distance to the bounding box is regressed by a linear layer $\mathcal{L}(\cdot)$:
\begin{equation}
    \hat{d}=\mathcal{L}(\mathcal{A}(\mathcal{F}(I), \mathcal{B})).
\end{equation}

However, in scenarios where the tracked object is small or when other objects are present in the image, relying solely on regression using the coarse box feature may not be an efficient approach. The inclusion of excessive background or interference from other objects can impede the network's performance and hinder the discovery of the effective object.

\noindent

\textbf{Point Feature Refiner.}
An alternative strategy is to utilize the point feature at the location of the labeled point from the whole image features $\mathcal{F}(I)$, denoted as $f_{point}$. 
A linear layer $\mathcal{L}(\cdot)$ is then applied to the point feature to regress the distance to the bounding box $\hat{d}$:
\begin{equation}
\hat{d}=\mathcal{L}(f_{point}),
\end{equation}
finally, the bounding box prediction is calculated by Eq.~\ref{equation: convert box}.

The problem of PFR is that only the features at the location of the labeled point are used for regression prediction, while the local features are struggling to capture the entire range of the object.

\subsection{Guided Click Refiner}
\label{TCR}

To further enhance the performance of the point-to-box refiner $\mathcal{R}$, we reassess the issues associated with the above strategies BFR and PFR, which can be summarized as follows:

(1) Lack of capability to input guiding information, as described in section \ref{intro}, requires the refiner to input guiding information in special scenarios to eliminate ambiguity caused by single-point input.

(2) The features used in the regression are either too holistic or local, leading to unstable effectiveness. Selecting the appropriate range of regression features can bring more stable and effective results.

(3) The structure of single-stage regression leads to low-quality outputs. 

Considering these factors, we introduce the Guided Convolution (GC), Prototype Selection module (PS), and Iterative Refinement module (IR), collectively forming the Guided Click Refiner (GCR).
The framework of GCR is illustrated in Fig.~\ref{fig:framework}. The GC utilizes guiding information to guide the regression process, enabling the capability to receive guiding information input. The Prototype Selection module aims to find suitable initial regression regions to exclude background or interference from other objects, thereby improving the regression quality. Finally, the Iterative Refinement module progressively optimizes the regression results through iterations, resulting in more accurate predictions.

\textbf{Guided Convolution.}
The Guided Convolution (GC) is the core structure within the Guided Click Refiner (GCR) model and plays a vital role in integrating guiding information with image features. Drawing inspiration from the concept of Dynamic Convolution~\cite{DBLP:conf/cvpr/ChenDLCYL20}, we believe that the parameters used for interacting with the image features should be generated based on the accompanying guiding information. This approach allows the guiding information to exert a significant guiding effect on the regression process. Based on this idea, we design the Guided Convolution (GC) structure.

The structure of GC is illustrated in Fig.~\ref{fig:framework}. GC takes the RoI feature $f_{roi}\in \mathbb{R}^{S \times S \times C}$ and guiding feature $f_{g}\in \mathbb{R}^{C}$ as inputs. 

It is important to note that there are two types of guiding features $f_{g}$, when it is necessary to specify the target category, the guiding information is text feature extracted from the natural language model Clip~\cite{radford2021learning}. When the target category does not need to be specified, the guiding information is a learnable proposal feature.

The process of GC structure is as follows: Firstly, the guiding feature $f_{g}\in \mathbb{R}^{C}$ is transformed into two sets of dynamic parameters $p_{1}\in \mathbb{R}^{C \times MC}$ and $p_{2}\in \mathbb{R}^{MC \times C}$, where $MC$ represents the middle channel and is a hyper-parameter. Simultaneously, the RoI feature $f_{roi}\in \mathbb{R}^{S \times S \times C}$ is reshaped to $f_{roi}^{'}\in \mathbb{R}^{(S \times S) \times C}$. Then, the interaction process between the dynamic parameters $p_{1}$, $p_{2}$ and $f_{roi}^{'}$ is as follows:
\begin{equation}
    f_{fusion} \in \mathbb{R}^{(S \times S) \times C} = (f_{roi}^{'} \times p_{1}) \times p_{2},
\end{equation}
where $\times$ represents matrix multiplication. The fusion feature $f_{fusion}^{'}\in \mathbb{R}^{C}$ can be obtained by passing $f_{fusion}$ through linear layer. The proposed Guided Convolution structure effectively utilizes guiding information to direct the RoI features, ensuring that the regression process yields the anticipated outcomes.

\noindent

\textbf{Prototype Selection.}
We designed the Prototype Selection (PS) module to select the most suitable initial regression range for each initial point. Following the idea from general object detectors, we set $k$ anchor boxes centered on the labeled point $p_0$ as the initial prototypes, where $k$ is the number of anchors ($m$ means scales, $n$ means aspect ratios, and $k=m \times n$). 

The Prototype Selection module consists of two consecutive Guided Convolution (GC) structures, each serving a distinct purpose. The first GC structure aims to perform initial adjustments on all prototype anchor boxes. This is necessary because the initially set anchor boxes may not fully cover all target scales. By making preliminary adjustments to the anchor boxes, they are brought as close as possible to the current target for regression. The second GC structure aims to predict the IoU (Intersection over Union) values for all adjusted anchor boxes. These predicted IoU values are then used to select the most suitable anchor box as the initial regression region. The process can be represented as follows:

For the first GC structure, we denote it as $GC_{1}(\cdot)$. Firstly, the RoI features $f_{roi}\in\mathbb{R}^{k\times S\times S \times C}$ corresponding to all prototype anchor boxes at the same point are extracted from image feature $\mathcal{F}(I)$ by RoIAlign. Simultaneously, a repeat operation transforms the guide feature $f_{g}\in\mathbb{R}^{C}$ into $f^{'}_{g}\in\mathbb{R}^{k \times C}$. Subsequently, both $f_{roi}$ and $f^{'}_{g}$ are fed into $GC_{1}(\cdot)$ to obtain fused features $f_{fusion}\in \mathbb{R}^{k\times (S\times S) \times C}$:
\begin{equation}
    f_{fusion}=GC_{1}(f_{roi}, f^{'}_{g}).
\end{equation}
Finally, the fused feature $f_{fusion}$ is transformed into the normalized distance $delta \in \mathbb{R}^{k \times 4}$ by linear layers. 

With $delta$, a new set of anchors, called refined anchors, is generated around the labeled point. The purpose of second GC structure $GC_{2}(\cdot)$ is to find the most suitable regression range from the refined anchors. We use IoU as the selection criterion. The process of predicting IoU in $GC_{2}(\cdot)$ is similar to the process of adjusting the initial anchor boxes in $GC_{1}(\cdot)$, with the following differences: (1) The RoI feature received by $GC_{2}(\cdot)$ is re-extracted from the image feature using the refined anchors. (2) The fused feature $f_{fusion}$ output by $GC_{2}(\cdot)$ is transformed into IoU prediction $S_{iou} \in \mathbb{R}^{k}$ through linear layers.
Finally, the anchor box with the highest IoU score is selected as the initial regression region.

\noindent

\textbf{Iterative Refinement.}
As mentioned above, the structure of single-step regression limits further improvement of the quality of the object bounding boxes. Therefore, we designed the Iterative Refinement (IR) module to enhance the quality of the object bounding boxes. The IR module adopts a cascaded structure by connecting multiple GC structures in series. It gradually refines the anchor boxes selected by the Prototype Selection module into higher-quality object bounding boxes. 

As shown in the Fig.~\ref{fig:framework}, the structure of the IR module is similar to that of the PS module, with the main differences being:
(1) For a single point, it only takes the initial regression region selected by the PS module as input.
(2) There is no structure for predicting IoU and only multiple refinements are applied to the received single anchor to obtain a higher-quality regression box.

\section{Experiment}
\label{sec:experiment}
\label{sec: TCR experiment}
This section focuses on the experiments of GCR. We first provide the implementation details of GCR and describe the benchmarks, including LaSOT~\cite{fan2019lasot} and GOT-10k~\cite{huang2019got}. Then, we present the results of GCR on these two benchmarks. Finally, we conduct ablation experiments to evaluate the effectiveness of crucial components in GCR and examine the impacts of essential parameter settings.
\subsection{Benchmark Details}
We primarily conduct experiments on the LaSOT~\cite{fan2019lasot} and GOT-10k~\cite{huang2019got}. The details of the benchmarks are as follows:

\textbf{LaSOT.}
is a large-scale, long-term tracking benchmark containing 280 videos with an average length of 2448 frames in the test set. 

\textbf{GOT-10k.}
is a large-scale benchmark covering a wide range of common challenges in object tracking.

\subsection{Implementation Details of \modulename}
Our codes are implemented based on MMTracking code-base~\cite{MMTracking_Contributors_OpenMMLab_Video_Perception_2021}, and the experiments are conducted on 8 NVIDIA RTX3090 GPUs.

\textbf{Model.}
We report the results of \modulename\ combined with STARK~\cite{yan2021learning} on LaSOT and GOT-10k benchmarks to evaluate the quality of boxes generated by \modulename. The default backbone of \modulename\ is ResNet-50~\cite{he2016deep}. The backbone is initialized with the parameters pre-trained on ImageNet~\cite{deng2009imagenet}. 

\textbf{Training.}
The default training schedule is 12 epochs. The optimizer is AdamW with the weight decay of 0.0001, and the initial learning rate is set to $10^{-4}$, divided by ten at epochs 8 and 11, respectively. It is worth noting that during the training process, to evaluate the effectiveness of \modulename\ with limited data, the experiments on the LaSOT and GOT-10k benchmarks are conducted using their respective training datasets. Each epoch randomly samples 64,000 images in the training dataset. 
The images are resized to $1333\times 800$. Graying, brightness and random horizontal flip are used for data augmentation. To enhance the efficiency of data utilization and improve the stability of the \modulename\ with different initial points, we adopt a method to generate random points by uniformly sampling within an ellipse during the training process. The ellipse is defined with semi-axes set to one-fourth of the width and height of the ground truth bounding box.

\subsection{Comparison of Different Initialization Methods}
 We compared GCR with various initialization methods. For a fair comparison, apart from the natural language trackers, the other initialization methods all use STARK-ST50 as the default tracker. As shown in Table~\ref{tab:different initial method.}, using ``Point + Text'' as the initialization method, GCR achieved a success rate of 65.0, significantly higher than BFR (53.2), GroundingDino~\cite{liu2023grounding} (57.8), OVSAM~\cite{yuan2024ovsam} (58.2) and PFR (60.3). Using ``Point" as the initialization method, GCR achieved a success rate of 62.4, higher than SAM-B (60.1), SAM-H (59.6), BFR (52.1), OVSAM (57.6) and PFR (60.5). This demonstrates GCR's higher accuracy in initializing single-object trackers through click interaction, proving the effectiveness of the GCR model. 

Additionally, we compared with ``Detector", ``Natural Language (NL)", and ``Natural Language + BBox" initialization methods. Compared to the highest-performing JointNLT, even with the ``NL + BBox" initialization method, GCR still achieved a significant advantage (65.0 vs 60.4). The advantage of GCR is even more evident when compared to initialization using only ``NL" (65.0 vs 56.9). 
Compared to the ``Detector (top1)" and ``Detector + Point" initialization methods, GCR still shows a significant advantage. The comprehensive comparison indicates the rationality of using click interaction as an initialization method for single-object trackers and the effectiveness of the GCR model.

\begin{table}[!ht]
    \centering
    \renewcommand{\arraystretch}{0.75}
    \setlength{\tabcolsep}{2.0pt}
    \begin{tabular}{cccccc}
    \toprule
         Initialize &Refiner & Tracker &Success &Precise &Norm-Precise \\
             \midrule
         \multirow{2}{*}{Detector (top.1)} & YOLO~\cite{jocher2021ultralytics} &STARK-ST50 &32.9 & 30.9 & 35.1\\
                                    & GLIP~\cite{li2022grounded} &STARK-ST50 &51.6 & 52.2 & 57.3\\
                                        \midrule
         \multirow{2}{*}{Dtector + Point}  & YOLO~\cite{jocher2021ultralytics} &STARK-ST50 &45.7 &45.4 &50.9  \\
                                    & GLIP~\cite{li2022grounded} &STARK-ST50 &58.5 &59.4 &65.3 \\
                                        \midrule
         \multirow{4}{*}{NL} &RTTNLD~\cite{DBLP:conf/wacv/FengABLS20} &- &28.0 & 28.0 & - \\
                             &TNL2K-1~\cite{DBLP:conf/cvpr/WangSZJW0W21} &- &51.0 & 49.0 & - \\
                             &CTRNLT~\cite{DBLP:conf/cvpr/LiYCP22} &- &52.0 &51.0 & - \\
                             &JointNLT~\cite{DBLP:journals/corr/abs-2303-12027} &- &56.9 &59.3 & - \\
                                 \midrule
         \multirow{4}{*}{NL + BBox} &RTTNLD~\cite{DBLP:conf/wacv/FengABLS20} &- &35.0 &35.0 & - \\
                             &TNL2K-2~\cite{DBLP:conf/cvpr/WangSZJW0W21} &- &51.0 &55.0 & - \\
                             &SNLT~\cite{DBLP:conf/cvpr/FengABS21} &- &54.0 &57.6 & - \\
                             &JointNLT~\cite{DBLP:journals/corr/abs-2303-12027} &- &60.4 &63.6 & - \\
                                 \midrule
        \multirow{6}{*}{Point} &BFR &STARK-ST50 &52.1 &52.7 &61.4 \\
                                &OVSAM~\citep{yuan2024ovsam} &STARK-ST50 &57.6 & 57.5 &64.0 \\
                                &SAM-H &STARK-ST50 &59.6 &61.1 &69.0 \\
                                &SAM-B &STARK-ST50 &60.1 &62.7 &70.7 \\
                                &PFR &STARK-ST50 &60.5 &62.6 &72.1 \\
                                \rowcolor{mygray}&GCR &STARK-ST50 &62.4 &65.1 &72.0 \\
                                    \midrule
        \multirow{5}{*}{Point + Text} &BFR &STARK-ST50  &53.2 & 53.1 &61.8 \\
                                &GroundingDino~\cite{liu2023grounding} &STARK-ST50 &57.8 &58.0 &64.0 \\
                                &OVSAM~\citep{yuan2024ovsam} &STARK-ST50 &58.2 &58.0 &64.5 \\
                                &PFR &STARK-ST50 &60.3 &62.0 &70.8 \\
                                \rowcolor{mygray} &GCR &STARK-ST50 &\textbf{65.0} &\textbf{68.3} &\textbf{75.7} \\
                                 \bottomrule
    \end{tabular}
    \caption{
    Comparison of different initialization methods. GLIP accepts the corresponding category as a prompt for detection, ``Detector (top 1)" refers to selecting the detection result with the highest score as the tracking target, ``Detector + Point" refers to selecting the detection result containing the initial point as the tracking target.
    }
    \label{tab:different initial method.}
\end{table}

\begin{table}[ht]
\renewcommand{\arraystretch}{0.75}
\centering
    \begin{center}
    \normalsize
    \begin{tabular}{cccccc}
    \toprule
        Initialize &Refiner & Tracker  & mAO & mSR$_{50}$ & mSR$_{75}$
          \\
        \hline
        \multirow{6}{*}{Point}   & \multirow{2}{*}{BFR}  
                                 & STARK-S50  &56.3 &62.1 &44.4\\
                                 & & STARK-ST50  &56.3 &62.1 &44.6\\ 
                                 \cline{2-6}
                                & \multirow{2}{*}{PFR} 
                                 & STARK-S50 &61.2 &69.7 &49.1\\
                                 & & STARK-ST50 &61.5 &69.9 &49.6\\ 
                                 \cline{2-6}
                                & \multirow{2}{*}{GCR} 
                                 & STARK-S50  &63.0 &71.3 &54.0 \\ 
                                &  &\cellcolor{mygray}STARK-ST50  &\cellcolor{mygray}\textbf{63.2} &\cellcolor{mygray}\textbf{71.4} &\cellcolor{mygray}\textbf{54.1}  \\ 
        \bottomrule
        \end{tabular}
         \caption{Performance comparisons of different refiners combined with different trackers on GOT-10k.}
    \label{tab:main result got10k}
    \end{center}
\end{table}

\begin{table}[h]
\renewcommand{\arraystretch}{0.75}
    \centering
    \setlength{\tabcolsep}{10pt}
    \begin{tabular}{lccccc}
    \toprule
    PS & IR & GC & ${\rm S}$ & ${\rm P}$  & ${\rm NP}$\\
    \hline
    \checkmark & & &61.1 &63.1 &71.5 \\
    \checkmark & & \checkmark &63.7 &66.5 &74.8 \\
     & \checkmark & &45.9 &45.5 &51.2 \\
      &\checkmark & \checkmark &50.3 &51.4 &56.8 \\
    \checkmark & \checkmark & &62.4 &65.1 &72.0 \\
    \checkmark & \checkmark & \checkmark &\textbf{65.0} &\textbf{68.3} &\textbf{75.7} \\
    \bottomrule
    \end{tabular}
    \caption{Different modules: PS (prototype selection), IR (iterative refinement) and GC (with text information). ${\rm NP}$ means norm precision evaluation matrix.}
    \label{tab:module ablation study}
\end{table}

\begin{table}[ht]
\renewcommand{\arraystretch}{0.75}
\setlength{\tabcolsep}{3pt}
    \centering
    \begin{tabular}{cccccc}
    \toprule
         Tracker &Initialize &Refiner &Success &Precise &Norm-Precise \\
         \midrule
        \multirow{7}{*}{MixFormer~\cite{cui2022mixformer}} &\multirow{4}{*}{Point}  &SAM-H~\cite{kirillov2023segment}  &59.8 &62.5 &69.3 \\
                                & &SAM-B~\cite{kirillov2023segment}  &60.6 &64.5 &71.1 \\
                                & &OVSAM~\cite{yuan2024ovsam}  &58.2 &59.2 &64.7 \\
                                & &\cellcolor{mygray}GCR  &\cellcolor{mygray}63.3 &\cellcolor{mygray}66.9 &\cellcolor{mygray}72.8 \\
                                \cline{2-6}
        &\multirow{3}{*}{Point + Text}  &GroundingDino~\cite{liu2023grounding}  &59.2 &65.8 &60.9 \\
                                & &OVSAM~\cite{yuan2024ovsam}  &58.6 &59.4 &64.6 \\
        &  &\cellcolor{mygray}GCR &\cellcolor{mygray}\textbf{66.4} &\cellcolor{mygray}\textbf{70.2} &\cellcolor{mygray}\textbf{76.5} \\
                                 \midrule
                \multirow{7}{*}{PrDimp~\cite{danelljan2020probabilistic}} &\multirow{4}{*}{Point}  &SAM-H~\cite{kirillov2023segment}  &50.7 &48.8 &57.6 \\
                                & &SAM-B~\cite{kirillov2023segment}  &51.9 &51.1 &59.8 \\
                                & &OVSAM~\cite{yuan2024ovsam}  &51.9 &49.2 &57.2 \\
                                & &\cellcolor{mygray}GCR  &\cellcolor{mygray}54.9 &\cellcolor{mygray}53.7 &\cellcolor{mygray}61.6 \\
                                \cline{2-6}
        &\multirow{3}{*}{Point + Text}  &GroundingDino~\cite{liu2023grounding}  &52.4 &49.7 &57.3 \\
                                & &OVSAM~\cite{yuan2024ovsam}  &51.3 &48.1 &55.6 \\
        &  &\cellcolor{mygray}GCR &\cellcolor{mygray}\textbf{58.4} &\cellcolor{mygray}\textbf{57.9} &\cellcolor{mygray}\textbf{65.9} \\
                                 \bottomrule
    \end{tabular}
    \caption{
    Comparison of different initialization methods with different trackers.
    }
    \label{tab:different trackers.}
\end{table}

\begin{table}[h]
\renewcommand{\arraystretch}{0.75}
    \centering
    \setlength{\tabcolsep}{18pt}
    \begin{tabular}{cccc}
    \toprule
    Stages & ${\rm S}$ & ${\rm P}$  & ${\rm NP}$\\
    \hline
    0 &63.7 &66.5 &74.8 \\ 
    2 &\textbf{65.0} &\textbf{68.3} &\textbf{75.7} \\ 
    4 &64.2 &68.0 &74.4 \\ 
    6 &64.7 &67.9 &74.9 \\
    \bottomrule
    \end{tabular}
    \caption{Ablation study of number of stages in iterative refinement.}
    \label{tab:number of stage ablation study}
\end{table}

\begin{table}[h]
\renewcommand{\arraystretch}{0.75}
    \centering
    \setlength{\tabcolsep}{1pt}
    \begin{tabular}{lccc}
    \toprule
    Settings & anchor scales & aspect ratios  & ${\rm S}$\\
    \hline
    \multirow{2}{*}{1 scale, 1 ratio}  &$128^{2}$                                    &1:1                &64.6 \\ 
                                       &$256^{2}$                                    &1:1                &64.4 \\
    \hline                                       
    \multirow{2}{*}{1 scale, 3 ratios} &$128^{2}$                                    &\{2:1, 1:1, 1:2\}  &64.6 \\
                                       &$256^{2}$                                    &\{2:1, 1:1, 1:2\}  &64.3 \\
    \hline
    4 scales, 3 ratios                  &\{$32^{2}$, $64^{2}$, $128^{2}$, $256^{2}$\} &\{2:1, 1:1, 1:2\}  &\textbf{65.0} \\
    \bottomrule
    \end{tabular}
    \caption{Ablation study of anchor setting in prototypes selection.}
    \label{tab:anchor setting ablation study}
\end{table}

\begin{table}[h]
\renewcommand{\arraystretch}{0.75}
    \centering
    \setlength{\tabcolsep}{16pt}
    \begin{tabular}{ccccc}
    \toprule
    $k$-times  & S & P  & NP   \\
    \hline
    1     &65.00   &68.10   &75.04 \\
    2     &65.03   &68.93   &75.30 \\
    3     &65.00   &69.06   &75.35 \\
    4     &64.91   &69.03   &75.19 \\
    5     &64.93   &69.00   &75.29 \\
    6     &64.90   &68.93   &75.26 \\
    \bottomrule
    \end{tabular}
    \caption{Robustness analysis of click position.}
    \label{tab:click position ablation study}
\end{table}

\begin{table}[]
\renewcommand{\arraystretch}{0.75}
\setlength{\tabcolsep}{3pt}
    \centering
    \begin{tabular}{cccccc}
    \toprule
        Number of Attempts & Initialize &Tracker &Success &Precise &Norm-Pricise  \\
        \midrule
        \multirow{2}{*}{Once} & Box &STARK-ST50 &54.1&51.9&60.9 \\
         & Point &STARK-ST50 &62.4&65.4&72.1 \\
        \midrule
        \multirow{2}{*}{Multiple} & Box &STARK-ST50 &55.7 &54.8 &63.8 \\
         & Point &STARK-ST50  &64.9 & 68.4 &74.9 \\
    \bottomrule
    \end{tabular}
    \caption{Single object tracking initialization simulation experiment. }
    \label{tab:simulation experiment}
\end{table}

\textbf{GOT-10k.}
On GOT-10k, we tested the performance comparison of GCR, BFR, and PFR using ``Point" as the initialization method. As shown in Table~\ref{tab:main result got10k}, combined with STARK-ST50, GCR achieved 63.2 mAO, while BFR and PFR achieved 56.3 and 61.5, respectively. The performance of GCR is better than both BFR and PFR, further demonstrating the effectiveness of our approach.

\subsection{Ablation Study}
\label{sec:ablation study}
To further analyze the effects of different modules in GCR, we conducted ablation experiments on LaSOT using STARK-ST50 as the default tracker.

\textbf{Modules.}
Ablation study of the Prototype Selection (PS),Iterative Regression (IR) and Guided Convolution (GC) modules are given in Table~\ref{tab:module ablation study}. When \modulename \ uses PS and IR modules separately, the success rate is 61.1 and 45.9. When these two modules are combined with GC to introduce text information into the regression process, the success rate increases by 2.6 (63.7 vs 61.1) and 4.4 (50.3 vs 45.9), respectively. When combining PS and IR modules, the success rate is 62.4. It is higher than the performance of the two modules alone and demonstrates the effectiveness of PS and IR module. When these two modules are combined with \modulename, the success rate increases again by 2.6 (65.0 vs 62.4). This demonstrates that GC effectively incorporates the textual information into the regression process.

\textbf{Different Trackers.}
We conducted comparative experiments on different initialization methods across various trackers. As shown in Table~\ref{tab:different trackers.}, the result indicates that GCR can provide more accurate initialization when combined with different trackers. It is worth noting that both MixFormer~\cite{cui2022mixformer} and PrDimp~\cite{danelljan2020probabilistic} are trackers with updating capabilities, which confirms that the impact of inaccurate initial box persists even in trackers with online updating capabilities.

\textbf{Number of Stages.}
The ablation study with different number of stages is shown in Table~\ref{tab:number of stage ablation study}. The result shows that the 2 stages achieve the best performance 65.0 success rate. When increasing the number of stages to 4 and 6, the success rate decreases slightly, but at the same time, there is a reduction in inference speed. Therefore, we select 2 stages as the default setting.

\textbf{Setting of Anchors.}
We conducted an ablation study of the different anchor settings. The result is shown in Table~\ref{tab:anchor setting ablation study}. The best performance is obtained with the 4 scales and 3 ratios setting, with success rate 65.0. Other anchor settings result in a slight decrease in success rate. Considering the diverse range of target sizes and aspect ratios in real-world scenarios, we chose the default setting of 4 scales and 3 ratios, even though it slightly increases the inference speed .

\subsection{Further Analysis}
To further analyze the effectiveness of GCR in real-time interactive tracking tasks, we conducted further analysis experiments of GCR.

\textbf{Robustness of Click Position.}
To validate the robustness of \modulename\ for the point annotation, we generate the random initial points several times during training process. Then, we test on LaSOT and average the final performance. As shown in Table~\ref{tab:click position ablation study}, in the case of averaging, the fluctuation in performance is negligible, further proving \modulename\ is robust to the initial point position.

\textbf{Simulation Experiments.}
To further verify the effectiveness of the GCR model, we conducted simulation experiments on the LaSOT dataset comparing box and point initialization. The experimental results are shown in Table~\ref{tab:simulation experiment}, where "Once" indicates that the operator only interacts once per video sequence, and "Multi" indicates that the operator is allowed to interact multiple times within a single video sequence to correct the initialization result. The experimental results further demonstrate that point input provides higher accuracy for single-object tracking initialization, and they also confirm the robustness of GCR to the input point location.

\textbf{Tracking Efficiency.}
We further validate the efficiency of \modulename\, which runs on one NVIDIA RTX3090 GPU and can reach 31 FPS. Meanwhile, the inference speed of STARK-ST50 is 26 FPS. Since the \modulename\ just generates a bounding box in the initial frame, \modulename\ only adds about 0.03 seconds to the inference time on each video sequence, which is a negligible increase. Therefore, when combined with the real-time trackers, the newly proposed ClickTrack paradigm still works well in real-time scenarios.

\begin{figure*}[htbp]
\centering
\includegraphics[scale=0.19]{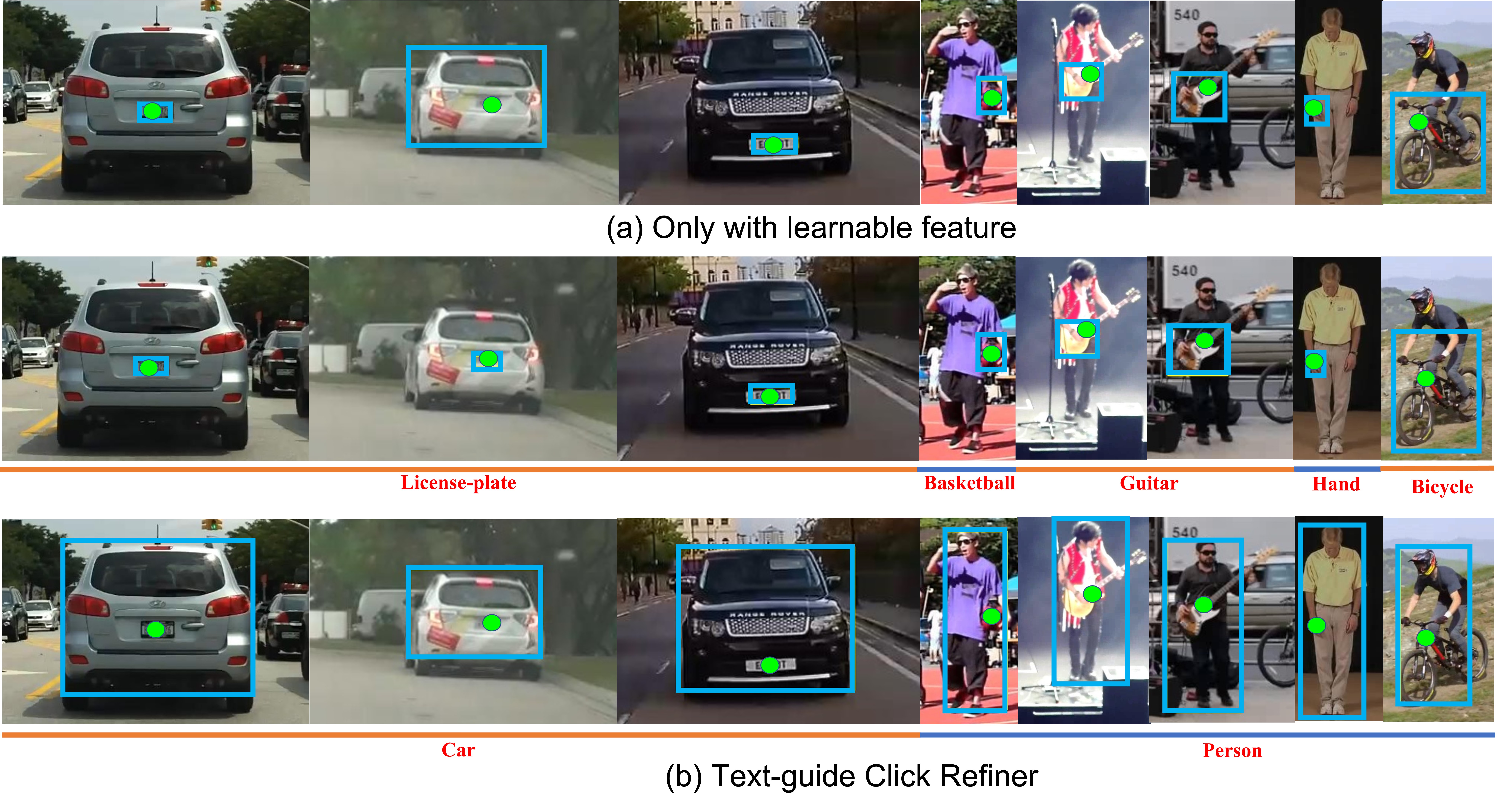}
\caption{Visualization of bounding box \textcolor{blue}{(blue)} generated with the same one point \textcolor{green}{(green)} and different text information \textcolor{red}{(red)}.}
\label{fig:Visualization of same point with different text}
\end{figure*}

\begin{figure}[tbp]
\centering
\includegraphics[width=0.8\linewidth]{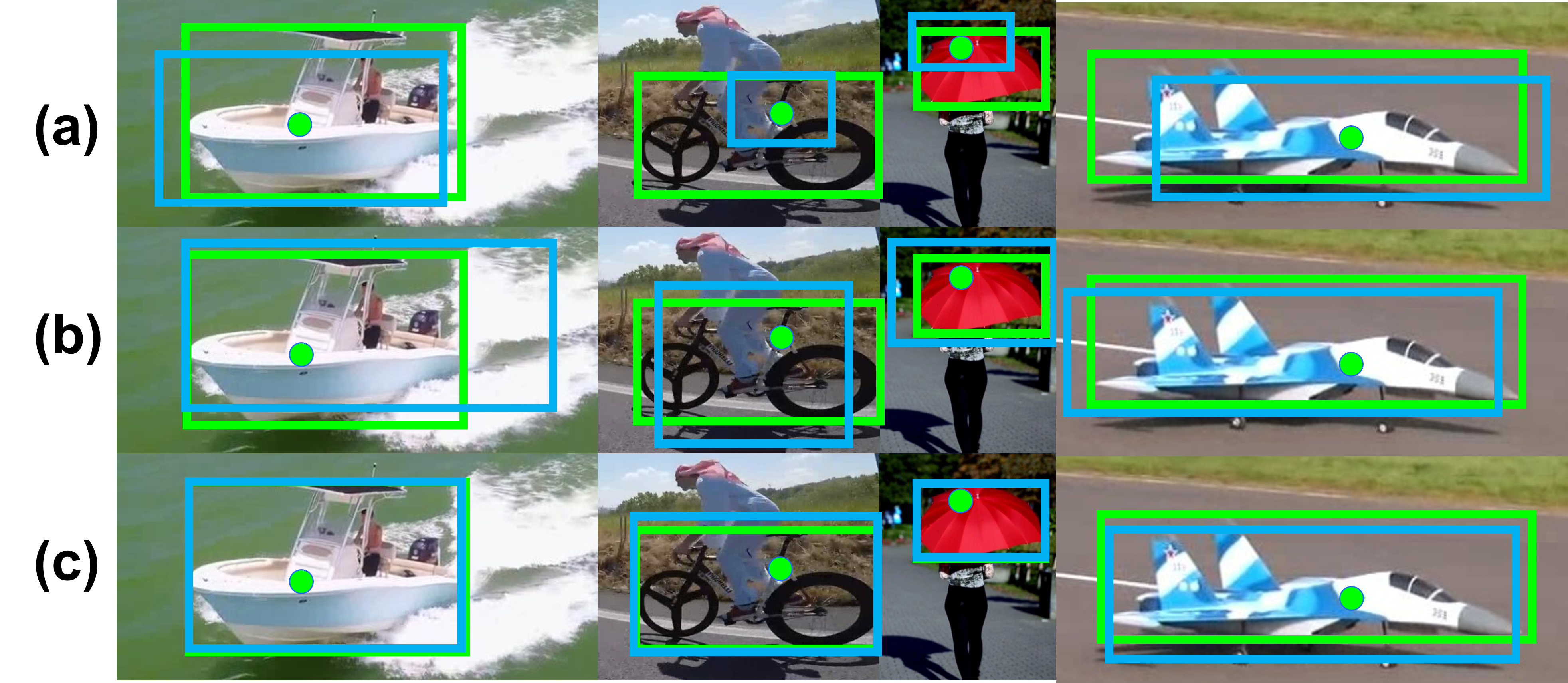}
\caption{Visualization of bounding box generated with different refiners. (a) BFR, (b) PFR and (c) \modulename. \textcolor{green}{Green} box is the ground truth and the \textcolor{blue}{blue} one is the box generated by different refiners.}
\label{reg_vis}
\end{figure}

\begin{figure}[tbp]
\centering
\includegraphics[width=0.9\linewidth]{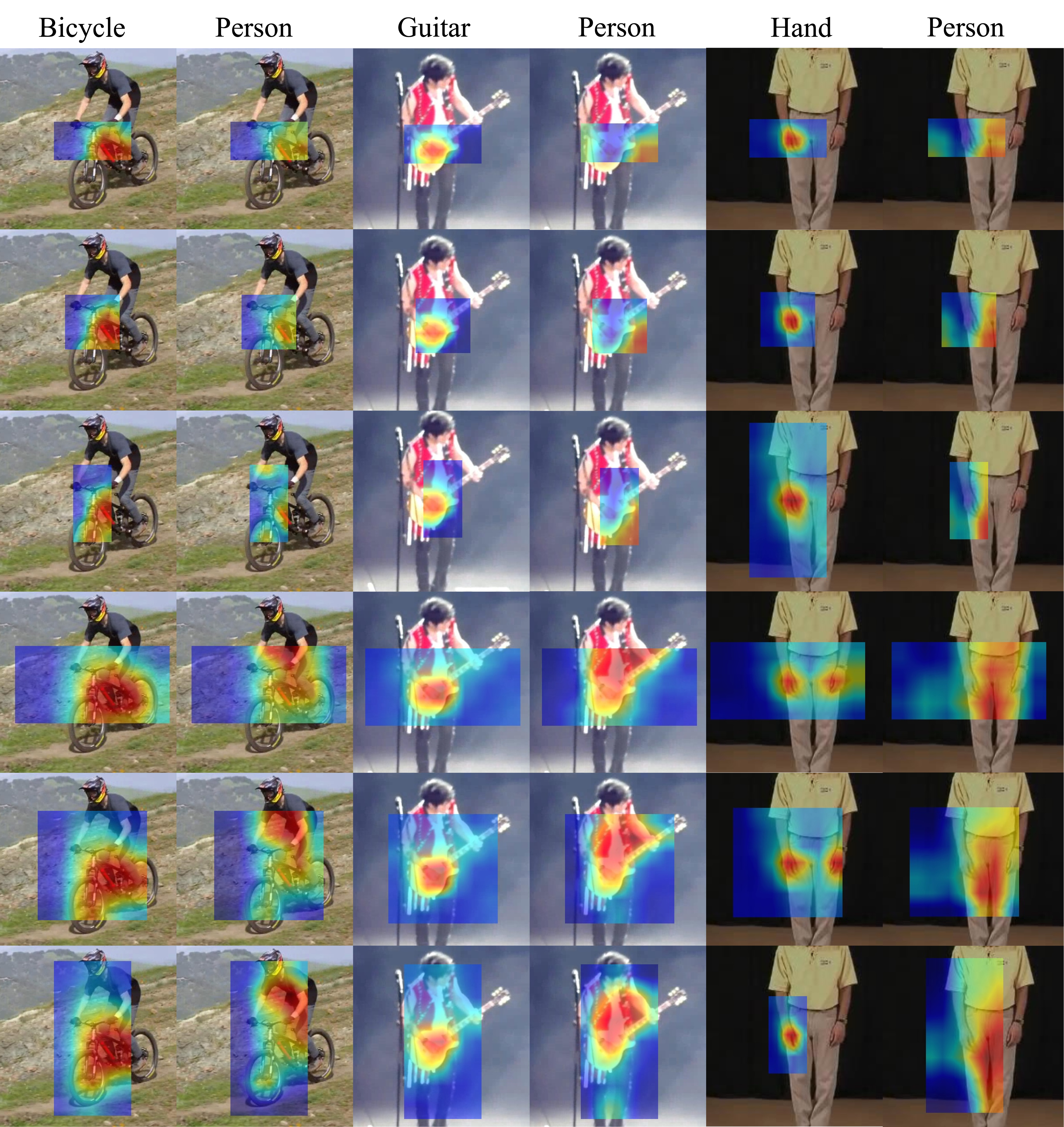}
\caption{Visualization of RoI features generated by GC structure with different text.}
\label{fig: heatmap_vis}
\end{figure}

\subsection{GCR visualization}
We select representative visualization results, as shown in Fig.~\ref{fig:Visualization of same point with different text}. The first row is the result of the learnable proposal feature, where it is impossible to control whether to produce partial or overall results. The second and third rows are the results generated by inputting different text information. Obviously, the text information can guide the network to produce the expected results. With different text information, \modulename\ can output the corresponding object target box. In Fig.~\ref{reg_vis}, we visualize the bounding box estimation of BFR, PFR and GCR. Our method further improves the bounding box quality.

Additionally, we used Eigen-CAM~\cite{muhammad2020eigen} to visualize the RoI features generated by the Guided Convolution (GC) structure with different textual feature. As shown in the Fig.~\ref{fig: heatmap_vis}, for clarity, we visualized a certain part of RoI features from the first GC module output in Prototype Selection. The visualization results demonstrate the effectiveness of Guided Convolution (GC) structure. Different text feature can make the RoI features focus on different area. We believe the main reasons for the alignment between visual features and textual features are: (1) The powerful zero-shot generalization capability of the CLIP model endows the textual features output by CLIP with inherent generalization ability. (2) In the GCR model, the textual features generated by CLIP are processed through an MLP layer. We believe that, through pre-training, the parameters of the MLP layer can further align the textual features with the visual features.

\section{Extending GCR to GCR-SAM}
\label{sec:TCR-SAM experiments}
\label{sec: TCR-SAM experiments}
In this section, we first explain the design motivation behind GCR-SAM, followed by a description of its overall structure. Finally, we present the main experimental results of GCR-SAM.
\subsection{Motivation}
Recently, the powerful segmentation ability of the visual large model Segment Anything Model (SAM) has gained widespread attention. SAM can accept three types of inputs, including the entire image, boxes and points. The input form of accepting points allows SAM to meet the task of VOS initialization in real-time interactive scenarios. However, we believe that SAM also faces ambiguity issues when receiving single-point inputs. In order to further explore the ability of the proposed GCR structure to eliminate the problem of single-point ambiguity, we combine the GCR head as a plugin with SAM, naming GCR-SAM. This combination empowers GCR with the strong segmentation ability of SAM, extending GCR's ability to produce mask-level results.

\begin{figure*}[htbp]
\centering
\includegraphics[scale=0.49]{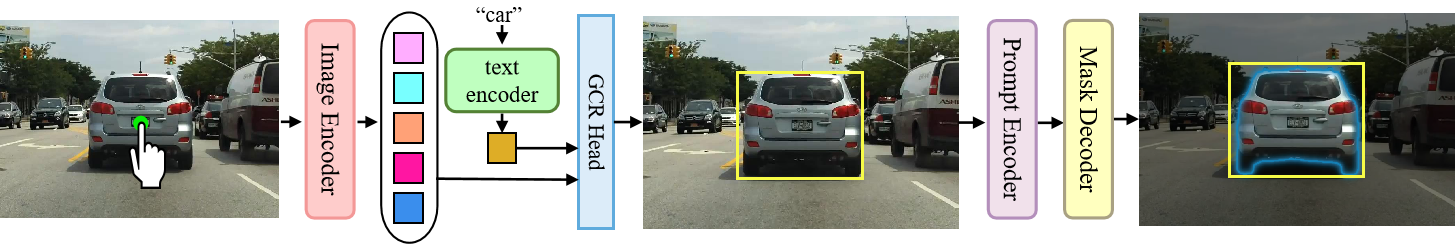}
\caption{The framework of TCR-SAM.}
\label{fig:tcr-SAM framework}
\end{figure*}

\subsection{The framework of GCR-SAM}
In this section, we present the comprehensive framework of GCR-SAM. As depicted in Fig.~\ref{fig:tcr-SAM framework}, the framework of GCR-SAM is straightforward. We retain the complete architecture of SAM, including the image encoder, prompt encoder, and mask decoder. During the training process, we keep SAM's pre-trained parameters frozen. Leveraging the non-ambiguity nature of SAM when accepting bounding boxes as prompt input, we integrated GCR's Prototype Selection (PS) and Iterative Regression (IR) modules as plugins into SAM. The two modules are represented as GCR head uniformly in Fig.~\ref{fig:tcr-SAM framework}. The GCR head can produce bounding boxes related to the input text, which serve as prompts for SAM, thereby generating mask results relevant to the input text. By integrating GCR with SAM, we successfully alleviate the ambiguity issues arising from single-point input.

\begin{table}[h]
\renewcommand{\arraystretch}{0.75}
    \centering
    \begin{tabular}{ccc}
    \toprule
         Method & average box IoU & average mask IoU
            \\
         \hline
         SAM         &48.4 &49.0   \\
         SAM-L &46.5 &51.2   \\
         SAM-M &49.8 &52.0   \\
         SAM-S &35.4 &37.9   \\
         GCR-SAM &72.9  &69.4 \\
         \rowcolor{mygray} GCR-SAM$^{*}$  &\textbf{74.4}  &\textbf{70.3} \\
        \bottomrule
    \end{tabular}
    \caption{The average IoU of GCR-SAM on COCO val 2017.}
    \label{tab:TCR-SAM coco}
\end{table} 

\begin{table*}[h]
\renewcommand{\arraystretch}{0.75}
\centering
    \begin{center}
        \setlength{\tabcolsep}{9pt}
            \resizebox{1.00\textwidth}{!}{
                \begin{tabular}{lccc|ccc|ccc}
                    \specialrule{0.085em}{0pt}{0pt}
                        \multirow{2}{*}{Initialize Method} & \multicolumn{3}{c|}{SiamMask~\cite{wang2019fast}} & \multicolumn{3}{c|}{FRTM~\cite{robinson2020learning}} &\multicolumn{3}{c}{UNINEXT-R50~\cite{yan2023universal}} \\
                            & $\mathcal{J\&F}$ & $\mathcal{J}$ & $\mathcal{F}$ 
                            & $\mathcal{J\&F}$ & $\mathcal{J}$ & $\mathcal{F}$
                            & $\mathcal{J\&F}$ & $\mathcal{J}$ & $\mathcal{F}$
                            \\
                        \hline
    Upper Bound  & 56.4 & 54.3 & 58.5 & 76.7 & 73.9 & 79.6 & 74.5 & 71.3 & 77.6  \\
    SAM    &31.6 &28.1 &35.2 &52.0 &48.3 &55.7               & 39.7 & 35.7 & 43.8  \\ 
    SAM-L    &48.9 &46.8 &50.9 &65.4 &62.6 & 68.1                             & 60.4 & 56.5 & 64.3  \\ 
    SAM-M    &34.7 &31.4 &38.1 &56.8 &52.9 &60.6                               & 44.0 & 40.2 & 47.8  \\ 
    SAM-S    &24.8 &20.9 &28.7 &42.5 &38.5 &46.5                               & 28.1 & 24.6 & 31.6  \\ 
    GCR-SAM       &54.7 &52.7 &56.7 &65.6 &62.2 &69.0 &62.6 &58.2 &67.0 \\
    \rowcolor{mygray} GCR-SAM$^{*}$ &\textbf{54.7} &\textbf{52.9} &\textbf{56.6} &\textbf{66.4} &\textbf{62.7} &\textbf{70.1} &\textbf{64.1} &\textbf{59.5} &\textbf{68.7} \\
         \specialrule{0.085em}{0pt}{0pt}
        \end{tabular}
        } 
    \end{center}
        
        \caption{Performance comparisons of different refiners combined with different VOS methods on DAVIS 2017 val dataset. ``Upper Bound" means using the precise mask as the initial for initialization.}
        \label{tab:main result DAVIS 2017 val}
    \label{tab:main}
\end{table*}

\begin{table*}[h]
\centering
    \begin{center}
    \setlength{\tabcolsep}{9pt}
\resizebox{\linewidth}{!}{
    \begin{tabular}{lccccc|ccccc|ccccc}
     \specialrule{0.13em}{0pt}{0pt}             
        \multirow{2}{*}{Initialize Method} & \multicolumn{5}{c|}{SiamMask~\cite{wang2019fast}} & \multicolumn{5}{c|}{FRTM~\cite{robinson2020learning}} &\multicolumn{5}{c}{UNINEXT-R50~\cite{yan2023universal}} \\
        & $\mathcal{G}$ & $\mathcal{J}_{s}$ & $\mathcal{F}_{s}$  & $\mathcal{J}_{u}$ & $\mathcal{F}_{u}$  
        & $\mathcal{G}$ & $\mathcal{J}_{s}$ & $\mathcal{F}_{s}$  & $\mathcal{J}_{u}$ & $\mathcal{F}_{u}$
        & $\mathcal{G}$ & $\mathcal{J}_{s}$ & $\mathcal{F}_{s}$  & $\mathcal{J}_{u}$ & $\mathcal{F}_{u}$
          \\
        \hline
    Upper Bound  &52.8 &60.2 &58.2 &45.1 &47.7 &72.1 &72.3 &76.2 &65.9 &74.1 &77.0 &76.8 &81.0 &70.8 &79.4   \\
    SAM   &41.3&42.1&43.9&37.2&42.2&52.8&50.4&54.6&49.5&56.7 &53.5 &48.3 &51.9 &53.1 &60.6\\ 
    SAM-L &31.5&38.6&38.6&23.4&25.4 &45.4&48.5&52.7&37.1&43.4 &45.1 &50.8 &55.5 &34.0 &40.0 \\
    SAM-M &42.0&43.9&45.7&37.2&41.2 &53.9&52.8&57.0&49.4&56.5 &54.9 &52.6 &56.4 &51.9 &58.5 \\
    SAM-S &31.4&28.5&32.4&29.4&35.4 &47.4&42.6&47.5&46.2&53.2 &42.9 &35.7 &39.8 &44.5 &51.6 \\
    GCR-SAM &44.2 &50.3 &50.0 &36.6 &39.8 &54.0 &55.7 &60.2 &46.9 &53.4 &57.3 &59.4 &63.9 &49.3 &56.4 \\
    \rowcolor{mygray} GCR-SAM$^{*}$ &\textbf{45.8} &\textbf{52.8} &\textbf{52.3} &\textbf{37.4} &\textbf{40.6} &\textbf{56.8} &\textbf{58.4} &\textbf{62.7} &\textbf{49.9} &\textbf{56.2} &\textbf{60.8} &\textbf{62.0} &\textbf{65.9} &\textbf{54.0} &\textbf{61.5} \\
         \specialrule{0.13em}{0pt}{0pt}
        \end{tabular}
        } 
        \end{center}
        \caption{Performance comparisons of different refiners combined with different VOS methods on YT-VOS 2018 val dataset. ``Upper Bound" means using the precise mask as the initial for initialization.}
        \label{tab:main result YT-VOS 2018 val}
\end{table*}

\begin{figure*}[htbp]
\centering
\includegraphics[scale=0.63]{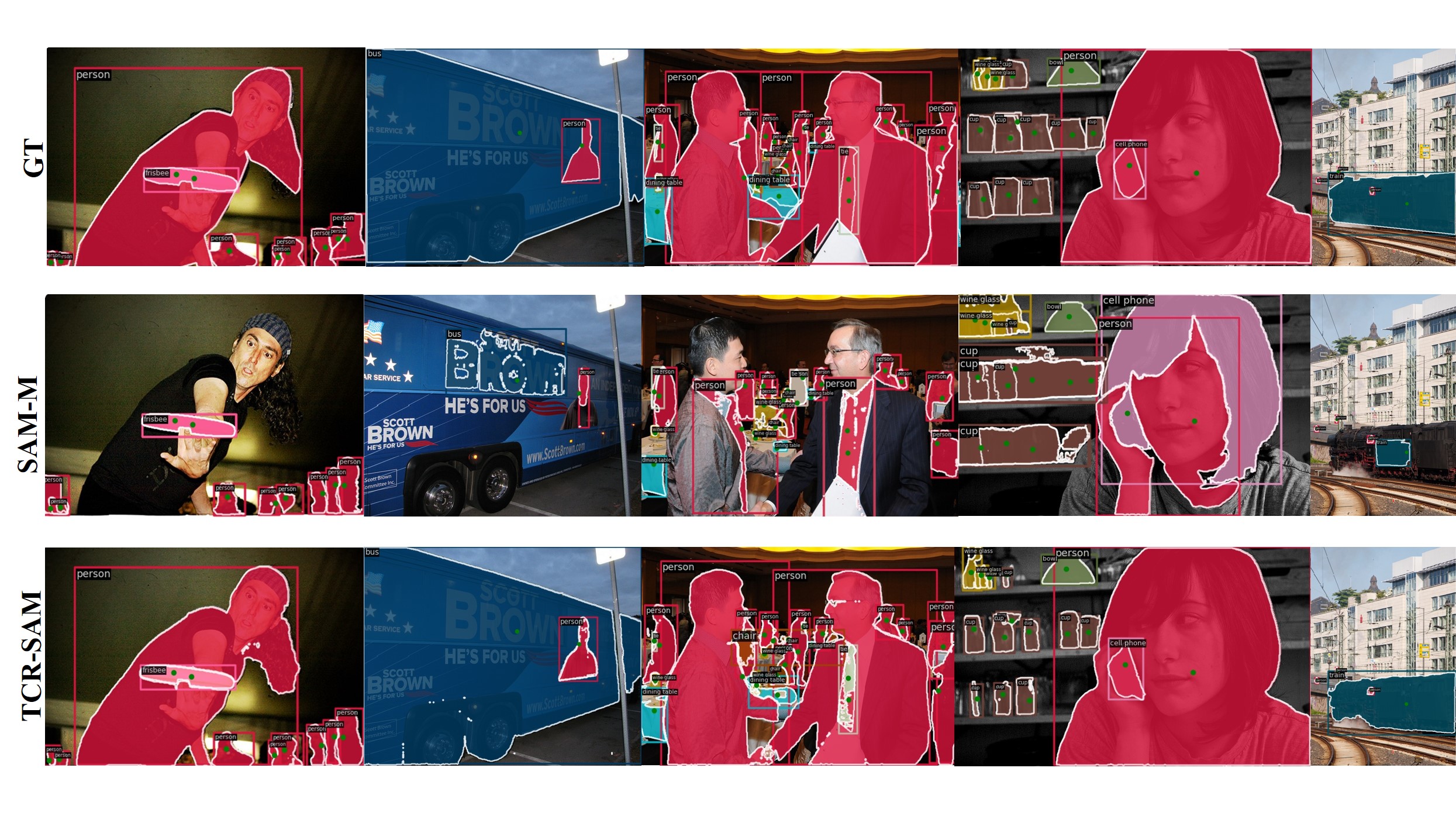}
\caption{Visualization of GCR-SAM. It is obvious that GCR-SAM effectively alleviates the ambiguity issues of single-point.}
\label{fig:Visualization of TCR-SAM}
\end{figure*}

\subsection{Main results of GCR-SAM}
In order to evaluate the quality of the masks generated by GCR-SAM, we conduct experiments on COCO~\cite{lin2014microsoft}, Davis 2017~\cite{pont20172017} and YT-VOS 2018~\cite{xu2018youtube}. On COCO, we select IoU as the evaluation metric. On DAVIS 2017 and YT-VOS 2018, we evaluate the masks quality by substituting the initial frame masks of three video object segmentation(VOS) methods, SiamMask, FRTM and UNINEXT-R50, with the mask generated by different methods. All experiments were conducted on the ViT-B SAM model. It is important to note that when SAM receives single-point inputs, it not only outputs the default mask but also generates masks of three different scales: large, medium, and small denoted as SAM-L, SAM-M and SAM-S respectively. In addition, we enhanced the performance GCR-SAM by constructing the training dataset which includes more categories, denoted as GCR-SAM$^{*}$. Due to computational resource limitations, we selected a subset of Objects365 (139,491 images, 284,967 instances) and combined it with COCO as the training dataset of GCR-SAM$^{*}$. 

DAVIS-2017 is a dataset for video object segmentation. It contains 150 videos - 60 for training, 30 for validation and, 60 for testing. Youtube-VOS 2018 is a Video Object Segmentation dataset that contains 4453 videos, 3471 for training, 474 for validation, and 508 for testing. The training and validation videos have pixel-level ground truth annotations for every 5th frame (6 fps). DAVIS-2017 adopts region similarity $\mathcal{J}$ , contour accuracy $\mathcal{F}$, and the averaged score $\mathcal{J\&F}$ as the metrics. Similarly, Youtube-VOS 2018 reports $\mathcal{J}$ and $\mathcal{F}$ for both seen and unseen categories, and the averaged overall score $\mathcal{G}$. It is worth noting that both datasets lack instance-level class annotations. Therefore, we manually annotated the category for each instance to meet the requirements of GCR-SAM for text inputs. 

\textbf{COCO.}
 As shown in Table~\ref{tab:TCR-SAM coco}, with the same point annotation as input, SAM(medium) obtains 49.8 average box IoU and 52.0 average mask IoU. It is the highest performance of SAM ViT-B model. In comparison, GCR-SAM obtains 72.9, 69.4 and GCR-SAM$^{*}$ obtains 74.4, 70.3 respectively. The significant improvement observed clearly demonstrates that GCR-SAM and GCR-SAM$^{*}$ can effectively produce results related to the input text, alleviating the ambiguity issues associated with single-point inputs. The visualization results of GCR-SAM in Section~\ref{sec: TCR-SAM Visualization} also indicate this conclusion. 

\textbf{DAVIS 2017.}
As shown in Table~\ref{tab:main result DAVIS 2017 val}, SiamMask, FRTM and UNINEXT-R50 obtain 56.4, 76.7 and 74.5 averaged score $\mathcal{J\&F}$ with precise mask. The masks with large scale generated by SAM obtain the best performance among SAM results, 48.9, 65.4 60.4 respectively. Combined with GCR-SAM, the three methods obtain 54.7, 65.6 and 62.6 averaged score $\mathcal{J\&F}$. Combined with GCR-SAM$^{*}$, the three methods obtain 54.7, 66.4 and 64.1 averaged score $\mathcal{J\&F}$. On the DAVIS dataset, SAM-L obtains the highest performance mainly because the validation set of DAVIS 2017 contains fewer testing instances (60 instances, 20 classes) and includes relatively simple instances. When using points as inputs, the ambiguity issue is practically non-existent.

\textbf{YT-VOS 2018.}
The comparisons of UNINEXT with different masks on Youtube-VOS 2018 are demonstrated in Table~\ref{tab:main result YT-VOS 2018 val}. SiamMask, FRTM and UNINEXT-R50 obtain 52.8, 72.1 and 77.0 averaged overall score $\mathcal{G}$ with the precise mask. The masks with medium scale generated by SAM obtain the best performance among SAM results, 42.0, 53.9 and 54.9 respectively. Combined with GCR-SAM, the three methods obtain 44.2, 54.0 and 57.3 averaged overall score $\mathcal{G}$. Compared to the DAVIS 2017 validation set, the YT-VOS 2018 validation set contains a larger number of instances and a more diverse range of classes (894 instances, 114 classes). Consequently, SAM-M obtains the highest performance. It aligns with the performance on COCO. The addition of more classes training data resulted in a significant improvement in GCR-SAM's performance, with GCR-SAM* achieving scores of 45.8, 56.8, and 60.8,  respectively.

The results on DAVIS and YouTube-VOS datasets demonstrate that GCR-SAM effectively alleviates the ambiguity issues present in SAM. It's worth noting that for experiments involving GCR-SAM on these two datasets, no training data from these datasets were used.

\textbf{GCR-SAM Visualization.}
\label{sec: TCR-SAM Visualization}
 We have selected some representative scenarios for visualization, as shown in Fig.~\ref{fig:Visualization of TCR-SAM}. We select the masks of SAM-M as the results for visualization because it obtained the highest IoU on COCO. The visual results clearly demonstrate the effectiveness of GCR-SAM in mitigating the ambiguity issue. 

\section{Conclusion}
In this paper, we rethink the SOT paradigm in real-time interactive scenarios and propose a new paradigm named ClickTrack, which aims to track arbitrary objects by clicking. We design the GCR model, which can convert the click point to a bounding box. GCR can be combined with arbitrary trackers to meet the needs of the real-time interactive scenario. Furthermore, we introduce text to GCR with a novel structure Guided Convolution (GC), which can greatly eliminate the ambiguity issue caused by the click interaction mode. Additionally, we combine GCR with SAM, and by introducing text, the ambiguity problem when SAM receives single-point input is greatly alleviated. Experimental results on multiple 
benchmarks demonstrate the effectiveness of GCR.

\appendix
\bibliographystyle{elsarticle-num}
\bibliography{bibliography}

\clearpage
\hspace*{\fill} 
\subsection*{  } %
\setlength\intextsep{0pt}
\begin{wrapfigure}{l}{25mm}
\centering
\includegraphics[width=1in,height=1.25in,clip,keepaspectratio]{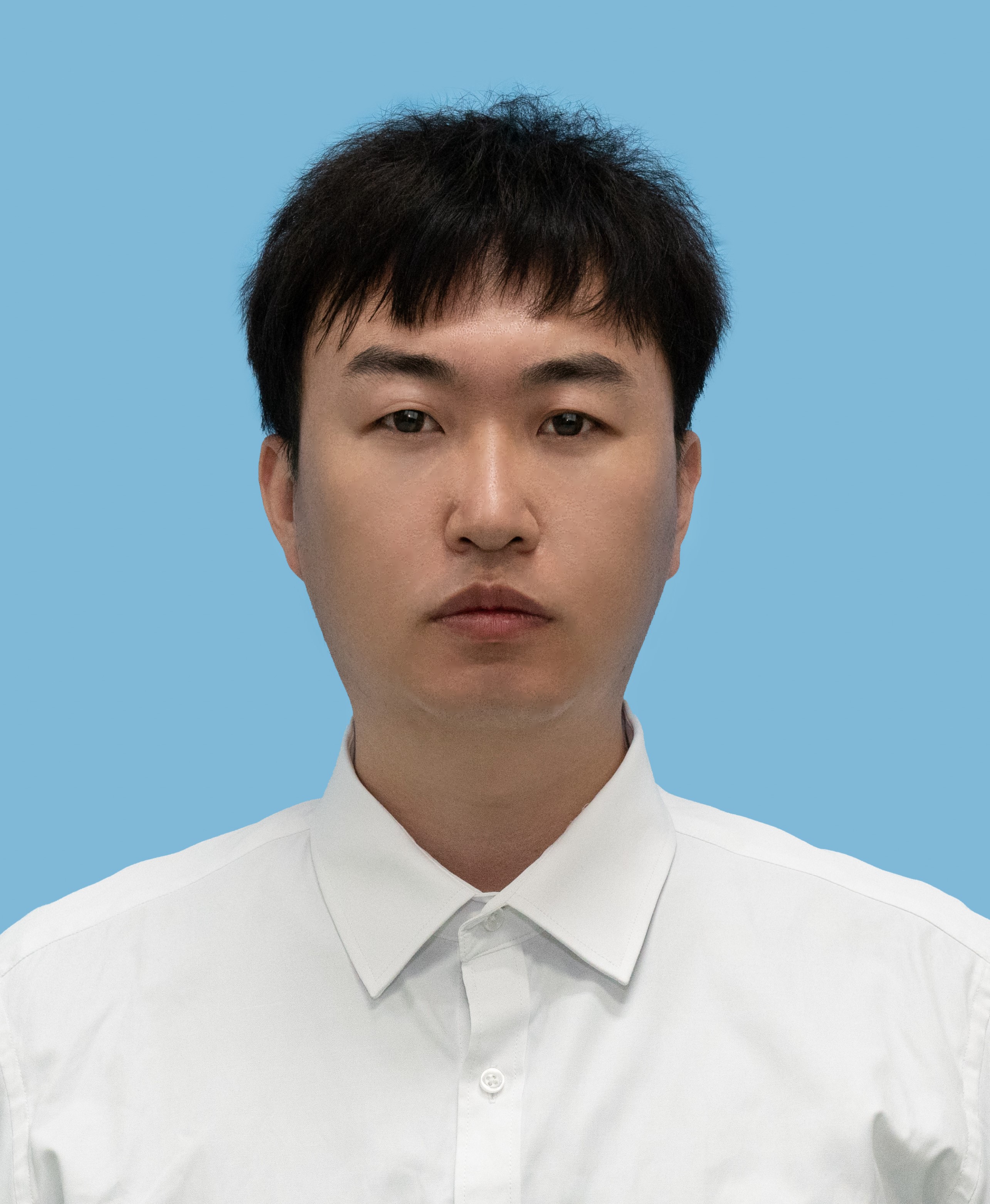}
\end{wrapfigure}\par
\noindent\textbf{Kuiran Wang} 
received the B.E. degree in computer science and technology from Central South University, China, in 2018. He is currently pursuing the Ph.D. degree in signal and information processing with University of Chinese Academy of Sciences. His research interests include machine learning and computer vision.\par

\hspace*{\fill} 
\subsection*{  } %
\setlength\intextsep{0pt}
\begin{wrapfigure}{l}{25mm}
\centering
\includegraphics[width=1in,height=1.25in,clip,keepaspectratio]{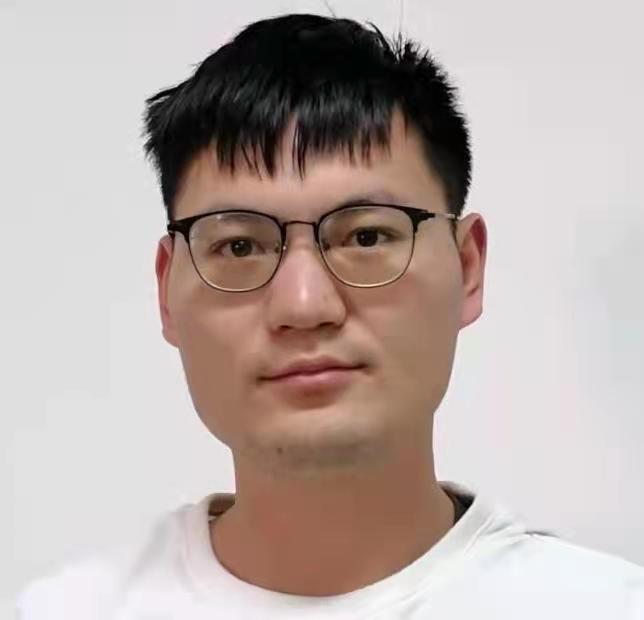}
\end{wrapfigure}\par
\noindent\textbf{Xuehui Yu} received the B.E. degree in software engineering from Tianjin University, China, in 2017. He is currently pursuing the Ph.D. degree in signal and information processing with University of Chinese Academy of Sciences. His research interests include machine learning and computer vision.\par

\hspace*{\fill} 
\subsection*{  } %
\setlength\intextsep{0pt}
\begin{wrapfigure}{l}{25mm}
\centering
\includegraphics[width=1in,height=1.25in,clip,keepaspectratio]{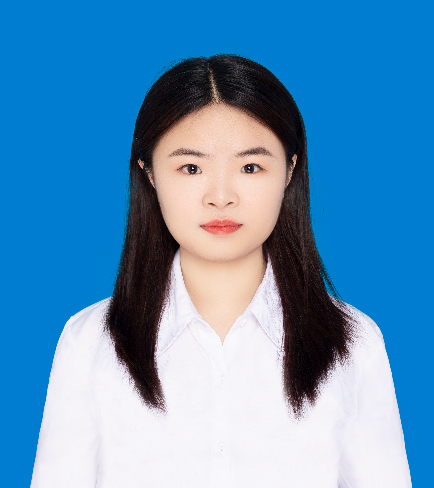}
\end{wrapfigure}\par
\noindent\textbf{Wenwen Yu} 
 received the B.E. degree in communication engineering from Wuhan University, China, in 2022. She is currently pursuing the M.S. degree in electronic and communication engineering with University of Chinese Academy of Sciences. Her research interests include machine learning and computer vision.\par

\hspace*{\fill} 
\subsection*{  } 
\setlength\intextsep{0pt}
\begin{wrapfigure}{l}{25mm} 
\centering
\includegraphics[width=1in,height=1.25in,clip,keepaspectratio]{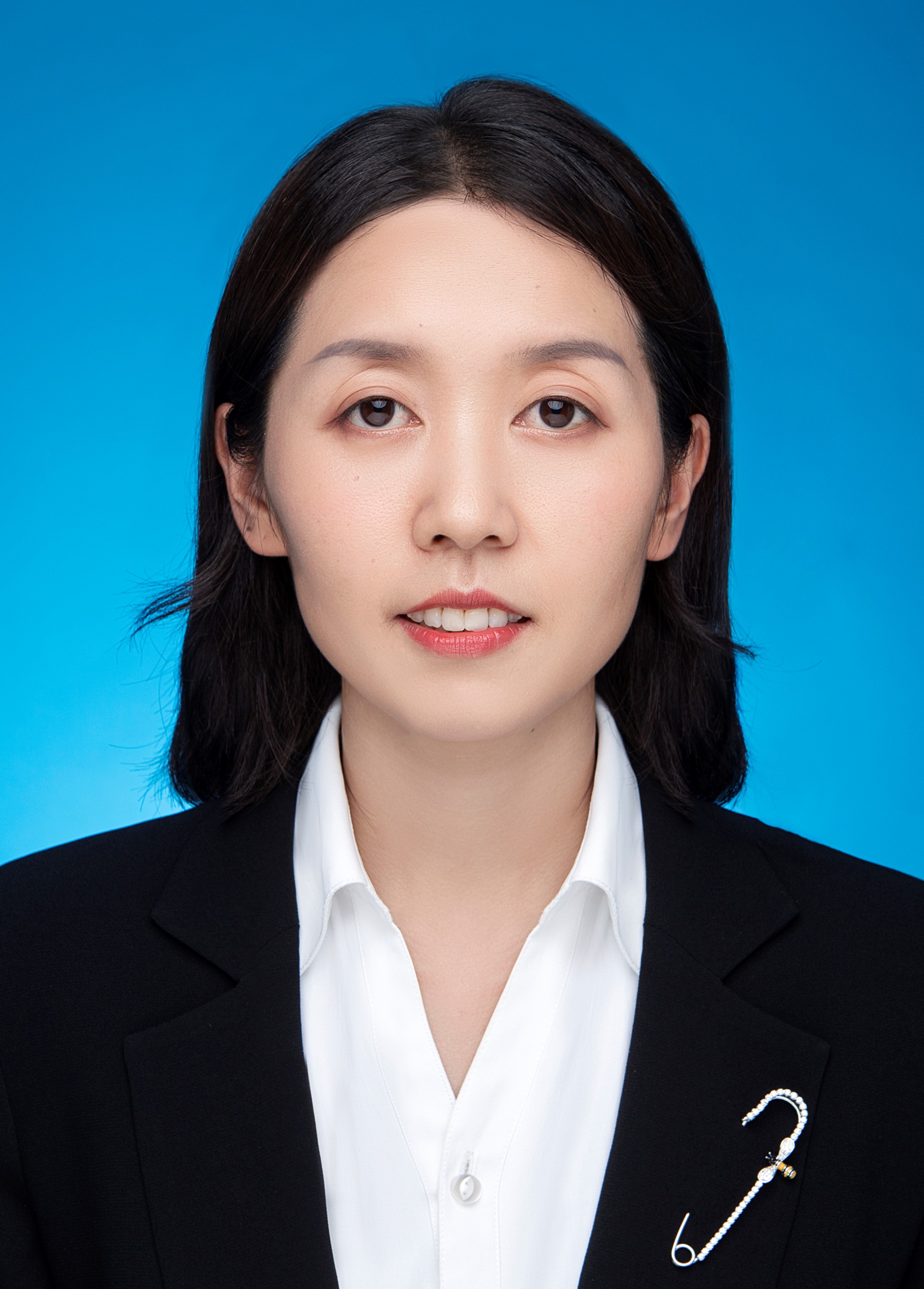}
\end{wrapfigure}\par
\noindent\textbf{Guorong Li}
received her B.S. degree in technology of computer application from Renmin University of China, in 2006 and Ph.D. degree in technology of computer application from the Graduate University of the Chinese Academy of Sciences in 2012. Now, she is an associate professor at the University of Chinese Academy of Sciences. Her research interests include object tracking, video analysis, pattern recognition, and cross-media analysis.\par

\hspace*{\fill} 
\subsection*{  } %
\setlength\intextsep{0pt}
\begin{wrapfigure}{l}{25mm}
\centering
\includegraphics[width=1in,height=1.25in,clip,keepaspectratio]{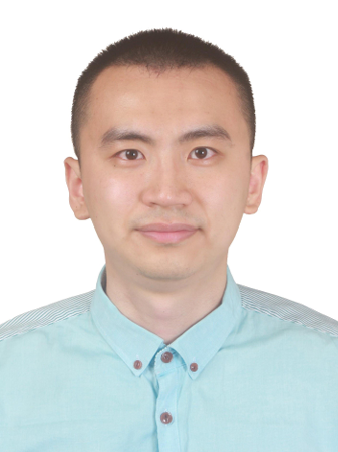}
\end{wrapfigure}\par
\noindent\textbf{Xiangyuan Lan} 
 received the B.Eng. degree in computer science and technology from the South China University of Technology, China, in 2012, and the Ph.D. degree from the Department of Computer cience, Hong Kong Baptist University, Hong Kong in 2016, where he is currently a Research Assistant Professor. His current research interests include intelligent video surveillance and biometric security.\par

\hspace*{\fill} 
\subsection*{  } 
\setlength\intextsep{0pt}
\begin{wrapfigure}{l}{25mm} 
\centering
\includegraphics[width=1in,height=1.25in,clip,keepaspectratio]{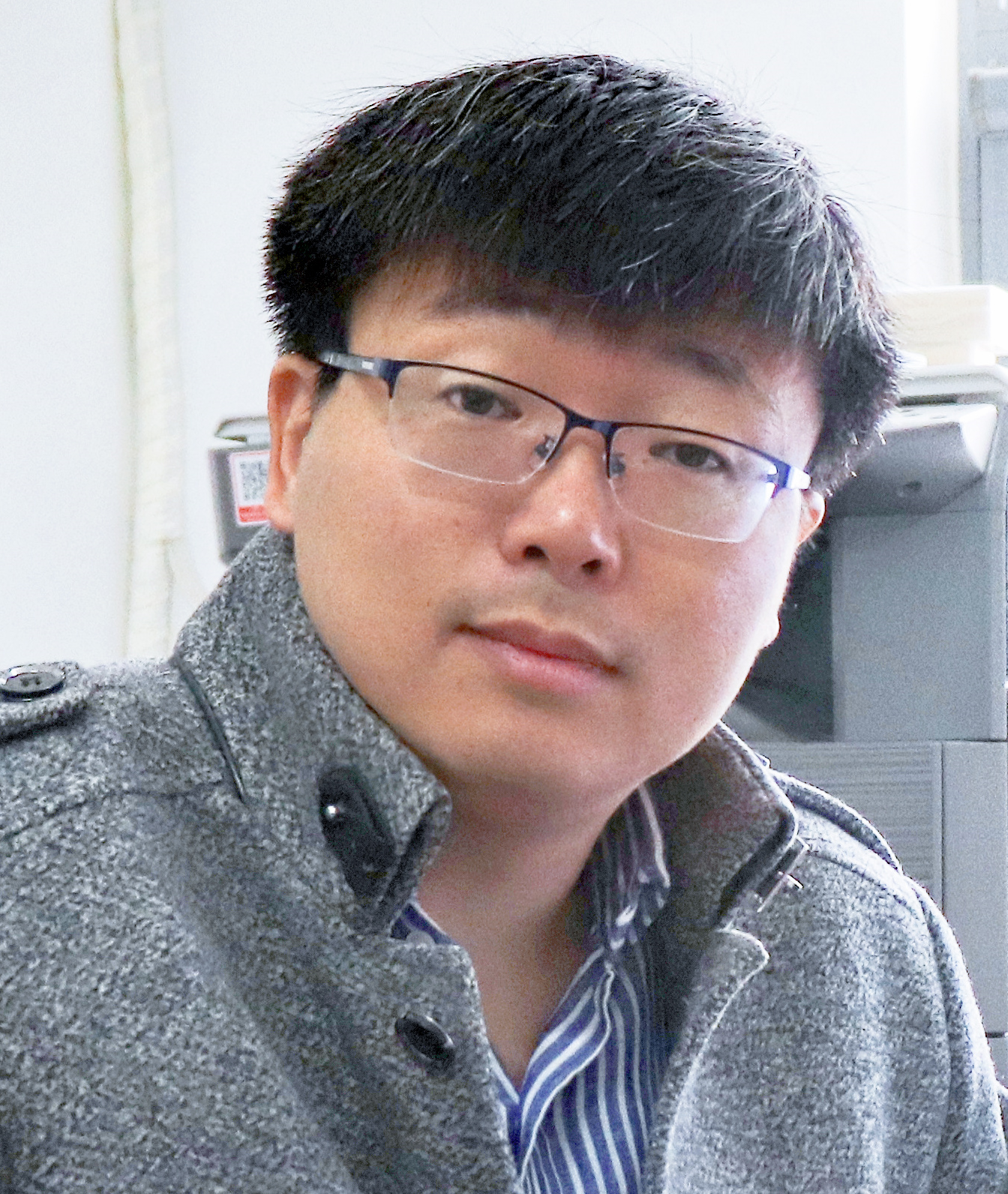}
\end{wrapfigure}\par
\noindent\textbf{Qixiang Ye} (M'10-SM'15) received the B.S. and M.S. degrees in mechanical and electrical engineering from Harbin Institute of Technology, China, in 1999 and 2001, respectively, and the Ph.D. degree from the Institute of Computing Technology, Chinese Academy of Sciences in 2006. He has been a professor with the University of Chinese Academy of Sciences since 2009, and was a visiting assistant professor with the Institute of Advanced Computer Studies (UMIACS), University of Maryland, College Park until 2013. His research interests include image processing, object detection and machine learning. He has published more than 100 papers in refereed conferences and journals including IEEE CVPR, ICCV, ECCV and PAMI.\par

\hspace*{\fill} 
\subsection*{  } 
\setlength\intextsep{0pt}
\begin{wrapfigure}{l}{25mm} 
\centering
\includegraphics[width=1in,height=1.25in,clip,keepaspectratio]{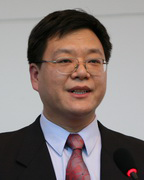}
\end{wrapfigure}\par
\textbf{Jianbin Jiao}
received the B.S., M.S., and
Ph.D. degrees in mechanical and electronic engineering
from Harbin Institute of Technology (HIT),
Harbin, China, in 1989, 1992, and 1995, respectively.
From 1997 to 2005, he was an Associate
Professor with HIT. Since 2006, he has been a
Professor with the School of Electronic, Electrical,
and Communication Engineering, University of the
Chinese Academy of Sciences, Beijing, China. His
current research interests include image processing,
pattern recognition, and intelligent surveillance.\par

\hspace*{\fill} 
\subsection*{  } 
\setlength\intextsep{0pt}
\begin{wrapfigure}{l}{25mm} 
\centering
\includegraphics[width=1in,height=1.25in,clip,keepaspectratio]{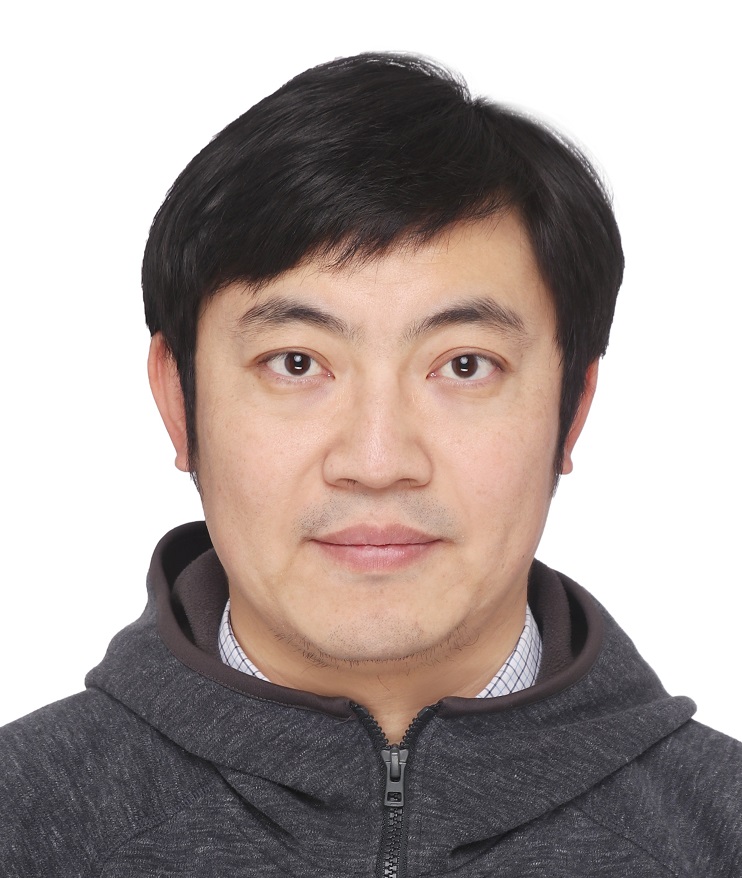}
\end{wrapfigure}\par
\noindent\textbf{Zhenjun Han}
 received the B.S. degree in software engineering from Tianjin University, Tianjin, China, in 2006 and the M.S. and Ph.D. degrees from University of Chinese Academy of Sciences, Beijing, China, in 2009 and 2012, respectively. Since 2013, he has been an Associate Professor with the School of Electronic, Electrical, and Communication Engineering, University of Chinese Academy of Sciences. His research interests include object tracking and detection.\par
\end{document}